%% file: ms.tex
\documentclass[12pt]{article}

\input{preamble.tex}

\title{Distributional Computational Graphs: Error Bounds}

\author[1]{Karl Olof Hallqvist Elias \thanks{olle.hallqvist.elias@signaloid.com}} 
\author[1]{Michael Selby}
\author[1]{Phillip Stanley-Marbell}
\affil[1]{Signaloid, 4 Station Square, Cambridge, CB1 2GE, United Kingdom}

\date{\today}

\begin{document}

\maketitle

\input{abstract.tex}

\keywords{ Wasserstein distances, Quantization of probability measures.}
\msc{Primary 60H10, 65C30; Secondary 60J60, 65C05}
\input{introduction.tex}
\input{notation-and-definitions.tex}
\input{core.tex}

\input{conclusions.tex}


\printbibliography

\end{document}

%% file: preamble.tex
\usepackage[utf8]{inputenc}
\usepackage[T1]{fontenc}
\usepackage[english]{babel}
\usepackage{csquotes}

\usepackage{amsmath, amssymb, amsthm, mathtools}
\usepackage{dsfont}

\usepackage{graphicx}
\usepackage{geometry}
\usepackage{caption}
\usepackage{subcaption}
\usepackage{float}
\usepackage{multirow}
\usepackage{array}
\usepackage{tikz}
\usetikzlibrary{shapes.geometric, arrows.meta, quotes,calc,positioning}

\usepackage{authblk}

\usepackage[style=numeric,sorting=none]{biblatex}
\usepackage{hyperref}

\usepackage{icomma}
\usepackage{xcolor}
\usepackage{comment}
\usepackage{todonotes}
\usepackage{algpseudocode}
\usepackage{algorithm}
\usepackage{xifthen}

\newboolean{displaywatermark}
\setboolean{displaywatermark}{false}  

\ifthenelse{\boolean{displaywatermark}}{%
\usepackage{draftwatermark}

  \SetWatermarkText{DRAFT}
  \SetWatermarkScale{4}
  \SetWatermarkColor{gray!20}
  \SetWatermarkAngle{45}
}{}

\newcommand{\keywords}[1]{\noindent\textbf{Keywords:} #1\par}
\newcommand{\msc}[1]{\noindent\textbf{MSC (2020):} #1\par}

\newcommand{\inedges}{\mathrm{in}}
\newcommand{\outedges}{\mathrm{out}}
\newcommand{\inbdry}{\partial_{\mathrm{in}}}
\newcommand{\depth}{\mathrm{depth}}
\newcommand{\Lip}{\mathrm{Lip}}

\newcommand{\indicator}{{\mathds 1}}

\newcommand{\id}{\mathrm{d}} 
\newcommand{\e}{\mathrm{e}}
\newcommand{\supp}{\mathrm{supp}\,} 
\newcommand{\diam}{\mathrm{diam}\, } 

\newcommand{\BR}{{\mathbb{R}}}
\newcommand{\BN}{{\mathbb{N}}}
\newcommand{\BP}{{\mathbb{P}}}
\newcommand{\BE}{{\mathbb{E}}}
\newcommand{\BZ}{{\mathbb{Z}}}

\newcommand{\CF}{{\mathcal{F}}}

\newcommand{\CM}{{\mathcal{M}}}
\newcommand{\CP}{{\mathcal{P}}}

\newcommand{\SFP}{{\mathsf{P}}}
\newcommand{\SFS}{{\mathsf{S}}}
\newcommand{\SFT}{{\mathsf{T}}}

\theoremstyle{plain}
\newtheorem{theorem}{Theorem}[section]

\newtheorem{lemma}{Lemma}[section]
\newtheorem{corollary}{Corollary}[section]

\theoremstyle{definition}
\newtheorem{definition}{Definition}[section]

\newtheorem{remark}{Remark}[section]
\newtheorem{assumption}{Assumption}

%% file: abstract.tex
%
%
\begin{abstract}
We study a general framework of distributional computational graphs: computational 
graphs whose inputs are probability distributions rather than point values. 
We analyze the discretization error that arises when these graphs are evaluated using finite 
approximations of continuous probability distributions. 
Such an approximation might be the result of representing a continuous real-valued distribution 
using a discrete representation or from constructing an empirical distribution from samples 
(or might be the output of another distributional computational graph).

We establish non-asymptotic error bounds in terms of the Wasserstein-1 distance, 
without imposing structural assumptions on the computational graph. 
\end{abstract}

%% file: introduction.tex
\section{Introduction}
Informally, a computer program is a sequence of arithmetic and logic operations that carry out a 
specific task. More abstractly, such a program is a composition of functions, where each function 
depends on the output of one or more preceding functions.
 A natural mathematical representation of this structure is a directed acyclic graph (DAG), in 
 which nodes correspond to intermediate values, and edges encode data dependencies and operations. 
 The use of DAGs in computer science is ubiquitous. They appear in compiler 
 theory~\cite[Chapter~9]{10.5555/286076}, neural network 
 models~\cite{blondel2025elementsdifferentiableprogramming,thost2021directed,SPERDUTI1997395,10.1007/3-540-44668-0_2}, 
 and probabilistic reasoning, notably in Bayesian networks and causal inference \cite{MR965765,10.5555/1795555}. 

This point of view, and in particular that of computational 
graphs~\cite[Chapter~4, Chapter~8.3]{blondel2025elementsdifferentiableprogramming}, is especially 
useful for modeling the structure of Monte Carlo programs. Such computations typically construct an 
approximation of a target probability distribution by generating a sequence of independent samples 
(random variates), which are then evaluated through a long sequence of arithmetic or functional 
operations \cite{petangoda2025montecarlomethodnew}. These operations, namely sampling, 
transformation, and aggregation, form a computational graph, where the structure of the graph 
captures the overall process.

Because of their simplicity of implementation and theoretical guarantees on convergence rates, Monte Carlo methods are widely used to approximate distributions not available in closed form. However, the convergence is slow. Suppose that the target distribution $\mu$ with cumulative distribution function (CDF) $F$ satisfies
\[
\int_\mathbb{R} \sqrt{F(t)(1 - F(t))} \, \mathrm{d}t < \infty.
\]
Then for $N$ independent samples, the expected Wasserstein-1 distance 
between the empirical and true distributions decays at rate $1/\sqrt{N}$ 
\cite[Theorem 1.1, Theorem 2.1]{MR1698999}.

This introduces a trade-off between 
computational cost and approximation fidelity. In addition the 
rescaled Wasserstein-1 distance between the target distribution $\mu$ and its empirical 
approximation $\mu_N$, with respective CDFs $F$ and $F_N$, satisfies the following CLT-type result \cite[Theorem 1.1]{MR1698999}:
\begin{equation*} 
\sqrt{N} \, W_1(\mu, \mu_{N}) \xrightarrow{d} \int_0^1 |B(t)| \, \mathrm{d}Q(t),
\end{equation*}
where \( Q : [0,1] \to \mathbb{R} \) is the quantile function of the target distribution and 
\( B(t) \) denotes a standard Brownian bridge. As a result, to quantify the random constant in the 
limit, one needs to understand the distribution of \( \int_0^1 |B(t)| \, \mathrm{d}Q(t) \). 
However, \( Q \) is only known approximately through \( F_N \), and moreover, the distribution of 
\( \int_0^1 |B(t)| \, \mathrm{d}Q(t) \) is not known in general. This necessitates a sampling 
procedure such as Monte Carlo. Thus, validating the results of a Monte Carlo simulation may itself 
require another layer of Monte Carlo. Furthermore, in practice, when relying on finite samples, a non-asymptotic analysis demonstrates a clear dependence on the dimension~\cite{MR4777182}.

Deterministic alternatives to Monte Carlo methods are typically based on quantization 
theory~\cite{MR1764176,MR4704055}, but are often tailored to specific applications, such as option 
pricing and optimal stopping problems~\cite{BallyPagesPrintemsAmerican,MR2046816}.

In light of these limitations, recent work \cite{9756254,10.1145/3466752.3480131,bilgin_patent} 
proposes a deterministic alternative to Monte Carlo estimation that operates by propagating 
approximations of the input distributions themselves through the computational graph. 
While this framework shares structural similarities with probabilistic graphical models 
\cite{10.5555/1795555,MR965765}, it addresses different problems. Probabilistic graphical models 
infer unknown distributions from data; distributional computational graphs (DCGs) propagate known 
distributions through deterministic operations. Because the distributions involved are typically 
continuous, prior research approximates them through quantization to enable this propagation. 
However, naively propagating quantized probability measures leads to exponential growth in 
computational and space complexity. Propagating $n$ quantized probability measures (via, e.g., 
convolution), each consisting of $N$ atoms, produces in the worst case a probability measure with 
$N^n$ atoms. This curse-of-dimensionality necessitates compression algorithms to maintain 
tractability. These algorithms reduce the space complexity of the resulting 
probability measures at a small cost in accuracy. With compression, this 
approach bypasses the need for sampling, removes the inherent stochastic variability, and offers 
theoretical guarantees under certain 
conditions~\cite{bilgin2025quantizationprobabilitydistributionsdivideandconquer}.

These prior research efforts focus on practical implementation strategies but 
do not present foundational theoretical insights into the strengths and 
limitations of the framework. We address this gap by providing rigorous error 
bounds that quantify how approximation errors propagate through the computational 
graph. Our theoretical analysis focuses on computational graphs that produce 
univariate output, though the framework is more broadly applicable. For algorithmic and implementation aspects, see Kirsten et al.~\cite{kirsten2025tensortrainapproachdeterministic}. We explore these further in a companion article. This includes algorithms for the proposed deterministic propagation, strategies to 
combat curse-of-dimensionality problems of this approach, and benchmarking against 
Monte Carlo simulations that extends the analysis 
of~\cite{bilgin2025quantizationprobabilitydistributionsdivideandconquer}.

The main contributions of this article are:
\begin{enumerate}
\item Rigorous error bounds for distributional computational graphs quantifying 
how approximation errors propagate to the output distribution.
\item A main result showing that output error is bounded by the complexity of the underlying graph (in a certain sense), a distortion factor determined by Lipschitz constants along paths, 
and the combined quantization and compression errors.
\item Application to the Euler-Maruyama scheme, deriving explicit bounds that 
combine discretization, quantization, and compression effects.
\end{enumerate}

The remainder of this article is organized as follows:
\begin{enumerate}
\item Section~\ref{s.not-def} introduces necessary concepts including the 
Wasserstein distance, quantization and compression algorithms, and the 
distributional computational graph framework.
\item Section~\ref{s.computational-graph-examples} provides concrete examples 
of computational graphs.
\item Section~\ref{s.upper-bound} contains the mathematical analysis and proofs 
of the main results.
\item Section~\ref{s:em-bound} proves Theorem~\ref{thm.em-bound} and establishes 
quantization error bounds for Gaussian distributions.
\item Section~\ref{section:conclusions} discusses assumptions, suggests 
improvements to the compression bound, and outlines future work.
\end{enumerate}

\subsection{Main result}
Let $G = (V, E)$ be a directed acyclic graph (DAG), where $(u,v) \in E$ denotes an edge going from $u$ to $v$. Assume that there exists a unique vertex $\Delta \in V$ with no outgoing edges with the additional property that every other vertex connects to $\Delta$ via a directed nearest-neighbor path. We define $\SFS \subset V$, referred to as the set of \emph{sources}, as the collection of vertices with no incoming edges. Let $\SFP(s, \Delta)$ denote the set of directed nearest-neighbor paths from the sources to the vertex $\Delta$.

Finally, to each node $v \in V \setminus \SFS$, associate a Lipschitz function $f_v : \BR^{\inedges(v)} \to \BR$, where $\inedges(v)$ denotes the set of incoming edges to $v$. Let $\CF =\left( f_v \right)_{v\in V}$ and for $v$ let $\|f_v\|_\Lip$ denote its Lipschitz constant. 

The output of the computational graph, denoted $x_\Delta$, is recursively computed via
\[
x_v = f_v\left( x_{\inedges(v)} \right), x_{\inedges(v)} \in \BR^{\inedges(v)}  \quad v \notin \SFS.
\]
Finally, define 
\[
F_\Delta : \BR^\SFS \mapsto \BR, x_S \mapsto x_\Delta
\]
as the function realized by the computational graph. The main theoretical result of this article is 
an upper bound on the Wasserstein-1 distance between the output of the true distributional 
computational graph and that of its compressed-and-quantized counterpart 
(see Definition~\ref{def.distributional-computational-graph}).

\begin{theorem}\label{thm.upper-bound}
Let $(G,\ \CF)$ be a computational graph and let $\mu_s, s \in \SFS$ be a collection of finite-mean probability measures. Let $\mu = \bigotimes_{s\in \SFS} \mu_s$, and let $\mu^{(n)} = \bigotimes_{s\in \SFS} \mu_s^{(n)}$ denote the quantized input measures. Furthermore, let $\mu_\Delta := \mu \circ F^{-1}_\Delta$ denote the output measure of the distributional computational graph and let $\mu^{(n),c}_\Delta$ denote the compressed-and-quantized output measure. Then, 
\begin{equation}
    W_{1}(\mu_\Delta,\mu_\Delta^{(n),c}) \leq 
    \sum_{s \in \SFS} 
    \sum_{\gamma \in \SFP(s,\Delta)}
        \left(
            W_1(\mu_s,\mu_s^{(n)})+\frac{3|\gamma|}{2^n}\diam(\supp(\mu_s^{(n)})) 
        \right)
        \prod_{(u,v) \in \gamma} \|f_v\|_\Lip.
\end{equation}    
\end{theorem}
We now make some comments on this result. 
\begin{remark}
Define 
\[
\mathsf{P}(\mathsf{S},\Delta) = 
\left\{
    \text{all directed paths from } \mathsf{S} \text{ to } \Delta
\right\},
\]
and let $\depth(G)$ be length of the longest path in $\mathsf{P}(\mathsf{S},\Delta)$.
Then we obtain the following cruder, but easier to interpret, upper bound:
\begin{equation}
\begin{split}
    W_{1}(\mu_\Delta,\mu_\Delta^{(n),c}) &\lesssim \depth(G) \cdot \# \mathsf{P}(\mathsf{S},\Delta) \cdot \max_{\gamma \in \mathsf{P}(\mathsf{S},\Delta) } \prod_{(u,v) \in \gamma} \|f_v\|_\Lip \\
& \quad \times  \max_{s \in \mathsf{S}} \left( W_1(\mu_s,\mu_s^{(n)}) + \frac{1}{2^n}\diam(\supp(\mu_s^{(n)})) \right).
\end{split}
\end{equation}
This demonstrates that the error is governed by four quantities:
\begin{enumerate}
\item The size of the underlying DAG, defined as $\depth(G) \cdot \# \mathsf{P}(\mathsf{S},\Delta)$,
\item The maximal distortion along all paths, $\max_{\gamma \in \mathsf{P}(\mathsf{S},\Delta) } \prod_{(u,v) \in \gamma} \|f_v\|_\Lip$,
\item The quantization error, $W_1(\mu_s,\mu_s^{(n)})$,
\item The compression error, $\frac{1}{2^n}\diam(\supp(\mu_s^{(n)}))$.
\end{enumerate}
In words:
\[
\mathrm{error}\lesssim
\mathrm{size\ of\ }G \times \mathrm{distortion\ factor} \times (\mathrm{quantization\ error} +\mathrm{compression\ error}).
\]
Without additional assumptions on the input distributions, the compression term 
is nearly sharp. We discuss this in more detail in Remark~\ref{rmk:compression}.
\end{remark}
As an extension of this result we apply the DCG framework to the Euler-Maruyama approximation scheme.
\begin{theorem}\label{thm.em-bound}
    Consider the Euler-Maruyama scheme \eqref{def.euler-maruyama}, with 
    $T>0, N\geq 1 , \Delta t := T/N$.
    Then under Assumptions \ref{a.sde-assumptions} and \ref{a.sde-symmetry}, there exists constants 
    $c=c(T,K)>0, c' = c'(T,K) >0$ such that 
    \begin{equation}\label{e.em-compressed-quantized}
    W_1\left(\mu_k,\mu_{k}^{(n),c}\right) \leq \frac{ c \e^{c' k \sqrt{n \Delta t}}}{2^n} 
    \end{equation}
    for all $k = 1,2,\dots,N,$ and $n \geq 0.$ 
\end{theorem}

The factor $\sqrt{n}$ in the exponent arises from the compression bound; we do not have numerical 
evidence to determine whether this is tight. Numerical simulations suggest that the term 
$k\sqrt{\Delta t}$ is sharp; see Remark~\ref{r:em-bound} for further discussion.

%% file: notation-and-definitions.tex
\section{Notation and Definitions}\label{s.not-def}
Given two functions $f,\ g :X \to \BR,$ we write $f(x) = \mathcal{O}(g(x))$ as $x \to x_0$ if 
\[
\limsup_{x \to x_0} f(x)/g(x)< \infty.
\]
Additionally, we write $f \sim g, x \to x_0$ if $\lim_{x \to x_0} f(x)/g(x)  = 1.$

Given a finite set $A$, we write $\mathbb{R}^A$ for the space of real-valued tuples indexed by $A$, 
so that $f(\{x_a : a \in A\})$ denotes evaluation without reference to a particular ordering 
of~$A$. If  $f: \BR^d \to \BR$ denotes a Lipschitz-continuous function, we let 
\begin{equation}\label{def.lipschitz-norm}
\|f\|_\Lip :=     \sup_{x,y : x\neq y} \frac{|f(x)-f(y)|}{\|x-y\|_1  }
\end{equation}
be the Lipschitz semi-norm where $\|x-y\|_1 = \sum_{i=1}^d |x_i-y_i|$ denotes the $\ell^1$-norm. We extend this to functions $f:\BR^A \to \BR$ in the canonical way.

Let $(\Omega,\ \CF)$ denote a measurable space. Given a set $A$ we define the indicator function of $A \subset \Omega$ by $\indicator_{A}$. For a probability measure $\mu$ on $(\BR,\ \CM)$ we let 
\begin{equation}
    F_\mu(x) := \mu(-\infty,x]
\end{equation}
denote the CDF. Furthermore, if $X$ is a random variable on $(\Omega,\ \CF)$ we write $X \sim \mu$ to mean $\BP(X \leq x) = F_\mu(x)$. If $A \subset X$ denotes a set of finite cardinality, we let  $\# A$ denote its cardinality. 

Lastly, a word about constants. We use $c, c', \dots $ to denote strictly positive constants whose values may vary throughout the article and even within the same display. When a constant depends on a model-specific quantity (say $\lambda$), we emphasize this by writing $c = c(\lambda)$.

\subsection{Directed acyclic graphs}
Given a directed acyclic graph (DAG), $G =(V,E)$, where $(u,v) \in E$ denotes the directed edge going from $u$ to $v$, we define the ingoing and outgoing edges by 
\begin{equation}
    \begin{split}
        \inedges(v) &= \{ u \in V : (u,v) \in E\},\\
        \outedges(v) &= \{ u \in V : (v,u) \in E\}.
    \end{split}
\end{equation}
Furthermore, given an edge $e\in E$ we let $e_-,e_+ \in V$ denote its tail and head, in other words
\(
e = (e_-,e_+).
\)
Given a set $A \subset V$ we let 
\[
\inbdry A = \{v \in A : \exists u \not \in A : (u,v) \in E \text{ or } \inedges(v) =\emptyset\}
\]
denote the in-boundary of $A$.

Moreover, we let $\SFS \subset V$, denote the set of sources:
\begin{equation}
    \SFS := 
    \left\{
        v \in V :     \# \inedges(v) = 0, \# \outedges(v)>0
    \right\},
\end{equation}
and note that $\SFS\neq \emptyset$ due to the fact that $G$ is acyclic. 

A nearest-neighbor path $\gamma $ between two vertices $u,v\in V$ is a finite sequence of edges 
$(u,v_1),(v_1,v_2),\ldots , (v_{n-1},v), \ v_i \in V$.
We let $\SFP(u,v)$ denote all directed nearest-neighbor paths 
between $u$ and $v,$ and we let $|\gamma|$ denote the length of $\gamma \in \SFP(u,v)$. 
Furthermore, we write $u\leq v$ (or $u\geq v$) if there exists a directed path from $u$ to $v$ 
(or from $v$ to $u$).
Finally, we let $d(u,v) = \inf\{ |\gamma| : \gamma \in \SFP(u,v)\}$ denote the graph distance 
between the vertices $u$ and $v.$ Note that $d$ does not define a metric since $d(u,v) \neq d(v,u)$.

Finally, we define the depth of $G$ by
\begin{equation}
\depth(G) 
:= \max
\left\{
    d(s,\Delta) : s \in \SFS
\right\}.
\end{equation}

\subsection{Wasserstein spaces}
Let $\CP_1$ denote the space of probability measures on $\BR$ with finite mean. 
For any two probability measures, $\mu,\ \nu \in \CP_1$, on $(\BR,|\cdot | )$ we define the Wasserstein metric \cite[Definition 6.1, p.105]{MR2459454} as
\begin{equation}
    W_1(\mu,\nu) = \inf \left\{ \BE[|X-Y|  ] : X \sim \mu, Y \sim \nu  \right\},
\end{equation}  
where the infimum is over all couplings between the probability measures $\mu$ and $\nu$ (or $X$ and $Y$).
Moreover, we have \cite[Theorem 3.1.2,p 109]{MR1619170}
\begin{equation}
    W_1(\mu,\nu) = \int_\BR \left | F_\mu(x)-F_\nu(x)\right | \id x.
\end{equation}
It is a well-known fact that Wasserstein distances metrize weak convergence \cite[Theorem 6.9]{MR2459454}.

\subsection{Quantization and compression algorithms}
Let $f: \CM \mapsto \BR$ be a continuous function with respect to the weak topology, satisfying $f(\mu) \in [\inf\supp(\mu),\sup \supp(\mu)]$. 
In this article we shall employ the following quantization algorithm, denoted 
$T^f : \CP_1 \times \BN \to \CP_1$, introduced by Bilgin et 
al.~\cite{bilgin2025quantizationprobabilitydistributionsdivideandconquer}.
Let $\Omega_\emptyset = \BR$ and define
\[
\Omega_- = \{x \in \Omega_\emptyset : x < f(\mu)\},\ \Omega_+ = \{x \in \Omega_\emptyset : x \geq f(\mu)\},
\]
and by assuming that  $\mu(\Omega_\pm)>0$ we also define
\[
\mu_\pm = \indicator_{\Omega_\pm} \mu / \mu(\Omega_\pm).
\]

We define the quantization algorithm by 
\begin{equation}\label{def.algorithm}
    \begin{split}
        T^f(\mu,n) 
        &:= \mu(\Omega_+)T^f(\mu_+,n-1)
        +\mu(\Omega_-)T^f(\mu_-,n-1), \\ 
        T^f(\mu,0) &:= \delta_{f(\mu)}.        
    \end{split}
\end{equation}
If $\mu(\Omega_\pm) = 0$, we define $T^{f}(\mu,n)=\delta_{f(\mu_\mp)},\ \forall n\geq 0$. 
Throughout the article, we shall frequently use the short-hand notation $\mu^{(n),f} := T^f(\mu,n)$.

We recall the following results obtained in \cite{bilgin2025quantizationprobabilitydistributionsdivideandconquer}.
\begin{lemma}[Lemma 3.1 in \cite{bilgin2025quantizationprobabilitydistributionsdivideandconquer}]\label{l.non-recursive}
    Let $n\in \BZ_{\geq 0}$ and let $\alpha \in \{\pm\}^n$.
    Define $\Omega_\pm,\mu_\pm$ as above and for $\alpha \in \{\pm\}^{n}, n\geq 1$ define $\Omega_{\alpha\pm}$ recursively by
    \[
    \Omega_{\alpha-} = \{x \in \Omega_\alpha : x < f(\mu_\alpha)\},\ \Omega_{\alpha+} = \{x \in \Omega_\alpha : x > f(\mu_\alpha)\}    
    \]
    and (assuming that $\mu(\Omega_{\alpha})>0$ for all $\alpha \in \{\pm\}^j, j = 1,2,\dots,n-1$)
    \[
    \mu_{\alpha\pm}  = \indicator_{\Omega_{\alpha\pm}} \mu / \mu(\Omega_{\alpha\pm}).
    \]
    Then
    \begin{equation*}\label{e.unravelled_full}
    \mu^{(n),f} = 
    \sum_{\alpha \in \{\pm\}^n} 
        \mu(\Omega_{\alpha}) \delta_{f(\mu_\alpha)},\  n \geq 0,
    \end{equation*}
    and
    \begin{equation*}\label{def.approx_cdf}
    F^{(n),f}(x) := \sum_{ \substack{ \alpha \in \{\pm\}^n : \\ f(\mu_\alpha) \leq x  } }   \mu(\Omega_{\alpha}).
\end{equation*}
\end{lemma}

\begin{lemma}[Lemma 4.3 in \cite{bilgin2025quantizationprobabilitydistributionsdivideandconquer}]\label{l.wasserstein-distance-approximation}
    \begin{equation*}
        W_1(\mu,\mu^{(n)}) = \sum_{\alpha \in \{\pm\}^n} \mu(\Omega_\alpha) W_1 ( \mu_\alpha , \delta_{f(\mu_\alpha)})
    \end{equation*}
\end{lemma}

For distributions $\mu$ such that $\mu(\Omega_\alpha)=0$ we can modify the results above in the following way. Let $\SFT_\mu^n \subset \bigcup_{j=0}^n\{\pm\}^j$ be a binary tree of height at most $n$ defined by 
\begin{equation*}\label{e.ttr-tree}    
\SFT_\mu^n  = 
\left\{
    \alpha \in \bigcup_{j = 0}^{n} \{\pm\}^j : \mu(\Omega_\alpha) >0 
\right\},
\end{equation*}
with the convention $\{\pm\}^0= \emptyset$. Let $\partial \SFT_\mu^n$ denote the set of leaves, then we instead obtain
\begin{equation*}\label{e.unravelled_partial}
    \mu^{(n),f} = T^f(\mu,n)=
    \sum_{ \alpha \in \partial \SFT_\mu^n}
        \mu(\Omega_{\alpha}) \delta_{f(\mu_\alpha)},\  n \geq 0,    
\end{equation*}
and 
\begin{equation*}
    W_1(\mu,\mu^{(n),f}) = \sum_{\alpha \in \partial \SFT_\mu^n} \mu(\Omega_\alpha) W_1 ( \mu_\alpha , \delta_{f(\mu_\alpha)}).
\end{equation*}
Given $X \sim \mu$ we say that $X^{(n) } \sim \mu^{(n)}$ is the quantized version of $X$.
We shall sometimes by abuse of notation write 
\[
T(X,n) = X^{(n)}.
\]

The proofs of Lemmas \ref{l.non-recursive} and \ref{l.wasserstein-distance-approximation} are easily generalized and so we state the following without proof.
\begin{lemma}
    For any $n\geq 0$ we have $X^{(n)} = T(X,n) = \sum_{\alpha \in \partial \SFT_\mu^n} 
    \indicator_{\{X \in \Omega_\alpha\}}\BE[X|X \in \Omega_\alpha]$ and $W_1(\mu,\mu^{(n)}) = \BE[|X-X^{(n)}|].$
\end{lemma}
In this article we shall restrict ourselves to the case when $f$ is the mean of the measure
\begin{equation}
    f(\mu) = \int_\BR t \id \mu(t),
\end{equation}
and so we drop the $f$ in the notation $T^{f}(\mu,n),\mu^{f,(n)}$ from now on.

We shall use the 
following~\cite[Eq. (32),p.16]{bilgin2025quantizationprobabilitydistributionsdivideandconquer} for 
our theoretical analysis. It states that for any discrete distribution
\[
\mu = \sum_{i=1}^m p_i\delta_{x_i}, \quad x_1 < x_2 < \dots < x_m,
\]
and any \( n = 1, 2, \dots, \lfloor \log_2 m \rfloor \), we have
\begin{equation} \label{e.discrete_upper_bound}
    W_1(\mu, \mu^{(n)}) \leq \frac{1}{2} \cdot \frac{\diam(\supp(\mu))}{2^{n}}.
\end{equation}
\begin{remark}\label{rmk:compression}
We remark that the upper bound of \eqref{e.discrete_upper_bound} is not sharp, as observed in previous work~\cite[Remark 4.3]{bilgin2025quantizationprobabilitydistributionsdivideandconquer}.
Indeed if $\mu =\sum_{j=1}^{2^m} \frac{1}{2^m} \delta_{j}$ for some $m \geq 0$, then by Definition~\ref{def.algorithm} we have
\begin{equation}
    \begin{split}
        W_1(\mu,\mu^{(n)})
        &= \frac{2^{m-n}}{4}\indicator_{\{m>n\}} = \frac{\diam(\supp(\mu))+1}{2^{n+2}}\indicator_{\{m>n\}} \\
        &\leq \frac{\diam(\supp(\mu))}{2^{n+1}}\indicator_{\{m>n\}}.
    \end{split}
\end{equation}
Furthermore, this shows that $\mu$ is a worst-case distribution: it nearly achieves the upper bound \eqref{e.discrete_upper_bound}, differing essentially only by the absence of a finite-size correction that would guarantee $W_1(\mu,\mu^{(n)}) = 0$ for all $n \geq m$. Also worth noting is that the collection of distributions $\left\{\mu =\sum_{j=1}^{2^m} \frac{1}{2^m} \delta_{j} : m \in \BN\right\}$ is not tight.

The bound \eqref{e.discrete_upper_bound} becomes particularly problematic in the case of quantized heavy-tailed distributions. Indeed if \( \mu \) denotes the quantized version of a Pareto-distributed random variable, the upper bound diverges as \( n \to \infty \) when the tail is sufficiently heavy~\cite[Example 4.2]{bilgin2025quantizationprobabilitydistributionsdivideandconquer}.

Finding a tighter bound for generic discrete distributions remains an open problem. We therefore 
use the existing bound in our analysis.
\end{remark}

\subsection{Distributional computational graphs}\label{s.computational-graph}
Formally we define a computational graph as the following. 

\begin{definition}\label{def.computational-graph}
Let $G = (V,\ E)$ be a directed acyclic graph (DAG) with the additional property that there exists a unique vertex $\Delta \in V$, \emph{the terminal vertex}, satisfying:
\begin{equation}
    \# \inedges(\Delta) > 0,\ \# \outedges(\Delta)=0,\ \Delta \geq v, \forall v \in V.
\end{equation}

We define a computational graph as the collection $(G, \CF)$, where $\CF = \{f : \BR^{\inedges(v)} \to \BR  : v \in V\setminus \SFS  \}$ is a collection of functions.
\end{definition}
Throughout the article we shall make the following assumption on the collection of functions.
\begin{assumption}
    We shall assume that for all $v \in V$, all functions $f_v \in \CF$ are globally Lipschitz, 
    \[
    |f_v(x)- f_v(y)|\leq \|f_v\|_\Lip \|x-y\|_1,\ x,y \in \BR^{\inedges(v)}
    \]
    where $\|\cdot\|_1$ is the $\ell^1$ norm on \(\BR^{\inedges(v)}\).
\end{assumption}

\begin{definition}
To each $s \in \SFS$ we associate a value $x_s \in \BR$  and we recursively define $x_v =f_v(\{x_u : u \in \inedges(v)\}) $ for $v \in V \setminus \SFS$. The value at the terminal vertex $\Delta$, i.e., 
\begin{equation}
    x_\Delta = f_\Delta(\{x_u : u \in \inedges(\Delta)\}),
\end{equation} 
is the \emph{output of the computational graph}.
\end{definition}

Finally we define the distributional (or random) counterparts.
\begin{definition}\label{def.distributional-computational-graph}
Given a computational graph $(G,\CF)$ with sources $\SFS$ and terminal vertex $\Delta$ and $n \in \{1,2,\ldots\}$:
\begin{itemize}
\item  A \emph{distributional computational graph} (DCG) is defined by replacing the deterministic input $x \in \BR^\SFS$ by a collection of independent random variables $(X_v)_{v \in \SFS}$. The output of the DCG is the random variable $X_\Delta$.
\item A \emph{quantized distributional computational graph} (qDCG) is a DCG where each input 
$X_s$ is replaced by its quantized version $X_s^{(n)} := T(X_s,n)$, where $T(\cdot,n)$ is the 
quantization operator from Equation~\ref{def.algorithm}.
\item A \emph{compressed-and-quantized distributional computational graph} (cqDCG) is a qDCG 
where we additionally compress intermediate distributions: for each $v \in V\setminus \SFS$, if 
$\# \supp(\mu_v^{(n)}) > 2^n$, we replace $X_v^{(n)}$ by $X_v^{(n),c} := T(X_v^{(n)}, n)$.
\end{itemize}
We let $(\mu_v)_{v \in V}$, $(\mu^{(n)}_v)_{v \in V}$, and $(\mu^{(n),c}_v)_{v \in V}$ denote the law of $(X_v)_{v \in V}$, $(X_v^{(n)})_{v \in V}$, and $(X_v^{(n),c})_{v \in V}$, respectively.
\end{definition}

\begin{remark}\label{r:cg-representation}
Note that the computational graph representation of a program is not unique. Indeed, consider the program:
\begin{algorithm}[H]
\caption*{{\bf Sum of $3$}}
\begin{algorithmic}
\State {\bf Input}: Three numbers $u,v,w$
    \State $x \gets u+v+w$
    \State {\bf Return} $x$
\end{algorithmic}
\end{algorithm}
A valid (though not realistic in practice) computational graph representation is to represent the computer program with the function $f(u,v,w) = u+v+w$.

\begin{figure}[H]
    \centering
    \begin{tikzpicture}[rounded corners]
    \begin{scope}[every node/.style={rectangle,thick,draw}]
        \node[draw] (u) at (-1.5,0) {$u$};
        \node[draw] (v) at (0,0) {$v$};
        \node[draw] (w) at (1.5,0) {$w$};
        \node[draw] (x) at (0,1.5) {$x = f(u,v,w)$};
    \end{scope}

    \begin{scope}[
                >={Stealth[black]},
                every node/.style={fill=white,rectangle},
                every edge/.style={draw,thick}
              ]
            \path [->] (u) edge   (x); 
            \path [->] (v) edge   (x);
            \path [->] (w) edge  (x); 
    \end{scope}
    \end{tikzpicture}
\end{figure}
Given that computers typically perform binary operations, a more realistic computational graph would define $h(u,v) := u+v$ and $g(x,w) := x+w$, with the composition $g(h(u,v), w)$ as the output:
\begin{figure}[H]
    \centering
    \begin{tikzpicture}[rounded corners]
    \begin{scope}[every node/.style={rectangle,thick,draw}]
        \node[draw] (u) at (-2,0) {$u$};
        \node[draw] (v) at (0,0) {$v$};
        \node[draw] (w) at (2,0) {$w$};
        \node[draw] (y) at (-1,1.5) {$y = u+v$};
        \node[draw] (x) at (0,3) {$x = g(y,w)$};
    \end{scope}

    \begin{scope}[
                >={Stealth[black]},
                every node/.style={fill=white,rectangle},
                every edge/.style={draw,thick}
              ]
            \path [->] (u) edge   (y); 
            \path [->] (v) edge   (y);
            \path [->] (y) edge   (x);
            \path [->] (w) edge   (x); 
    \end{scope}
    \end{tikzpicture}
\end{figure}
From this perspective one could assume 
\[\# \inedges(v) \leq 2, \ \forall v \in V.\] 
In our analysis this assumption does not improve any of our results and so we chose not to impose 
this constraint. 
It is reasonable to expect, however, that with additional assumptions on the computational graphs 
and distributions, this constraint could affect the quality of the bounds. We discuss this assumption further in Section~\ref{section:conclusions}.
\end{remark}
We end the section by briefly explaining how Monte Carlo methods fit into the computational graph framework. Given a computational graph \( (G, \mathcal{F}) \) with terminal vertex \( \Delta \) and source vertices \( \SFS \), the goal is to approximate the distribution \( \mu_\Delta \) at the terminal vertex. To do so, one generates \( N\geq 1 \) independent samples of the source vector \( (X_s)_{s \in \SFS} \), denoted by \( (X_s^i)_{s \in \SFS} \) for \( i = 1, \dots, N \). Monte Carlo simulation then propagate these inputs through the computational graph, producing corresponding outputs \( X_\Delta^i \) at the terminal vertex.

The underlying computer program that defines the graph typically implements this process automatically. The empirical distribution at \( \Delta \) is then defined as
\begin{equation}
    F_\Delta^N(t) := \frac{1}{N} \sum_{i=1}^N \mathbf{1}\{X_\Delta^i \leq t\},
\end{equation}
which serves as an approximation to the true cumulative distribution function \( F_\Delta \) of \( \mu_\Delta \).

\subsection{Examples}\label{s.computational-graph-examples}
This section provides two examples of (distributional) computational graphs. The first is the Euler-Maruyama approximation scheme for stochastic differential equations (SDEs) and the second is a simple sorting algorithm that produces the order statistic of a collection of independent random variables.
\subsubsection{The Euler-Maruyama approximation scheme}
Let $(W_t)_ {t \geq 0 }$ be a be Brownian motion defined on a complete filtered probability space $(\Omega, \CF, (\CF_t)_{t \geq 0 },\BP)$. We let $(X_t)_{t \geq 0 }$ denote the stochastic process defined by the stochastic differential equation (SDE) 
\[
\id X_t  = a(t,X_t) \id t + b(t,X_t) \id W_t .
\]
We shall moreover use the standard (global) assumptions on $a,b$ that guarantee existence and uniqueness of the solution, see \cite[p.128]{MR1214374}.
\begin{assumption}\label{a.sde-assumptions}
    Let $T>0$.
\begin{enumerate}
    \item 
    $a,b : [0,T] \times \BR \to \BR$ are jointly measurable in $(t,x).$ 
    \item There exists a constant $K = K(T)>0$ such that 
    \begin{equation}
        \begin{split}
            &|a(t,x)-a(t,y)| \leq K |x-y| , \\
            &|b(t,x)-b(t,y)| \leq K |x-y| , \\
        \end{split}
    \end{equation}
    for all $t \in [0,T], x,y \in \BR.$
    \item There exists a constant $K'= K'(T)>0$ such that 
    \begin{equation}
        \begin{split}
            &|a(t,x)| \leq K' ( 1 +|x|) , \\
            &|b(t,x)| \leq K' ( 1 +|x|) . \\
        \end{split}
    \end{equation}
    \item $X_0$ is $\CF_0$ measurable with $\BE[|X_0|]<\infty.$
\end{enumerate}
\end{assumption}
We now outline the Euler-Maruyama scheme. Given the interval $[0,T]$ we shall consider a 
discretization with equal step-size. In other words, let $N\geq 1$ be an integer and define the 
discretization by $t_j = jT/N, j = 0,1,\dots,N$ and note that $t_j-t_{j-1} = T/N =: \Delta t$.

The classical Euler-Maruyama scheme, \cite[Equation (1.3), p.304]{MR1214374} is: 
\begin{equation}\label{def.euler-maruyama}
    Y_{k+1} =   Y_{k} + a(t_k,Y_{k}) \Delta t + b(t_k,Y_{k}) \sqrt{\Delta t}  \xi_k,\ k = 0,1,2,\dots,N-1
\end{equation}
where $\xi_k\sim \mathcal{N}(0,1)$ are i.i.d and $Y_{0} = X_0.$ To phrase this as a computational graph we define the functions 
\begin{equation}\label{e.em-functions}
    \begin{split}
    f_{k,1}(y) &:= y+a(t_k,y)\Delta t,\\
    f_{k,2}(\xi,y) &:= \sqrt{\Delta t}b(t_k,y) \xi,\\
    f_{k,3}(a,b) &:= a+b.    
    \end{split}    
\end{equation}
so that 
\begin{equation*}
    Y_{k+1} = f_{k,3}(f_{k,1}(Y_k), f_{k,2}(\xi_k,Y_k)).
\end{equation*}
With these definitions, we can illustrate the computational graph as:
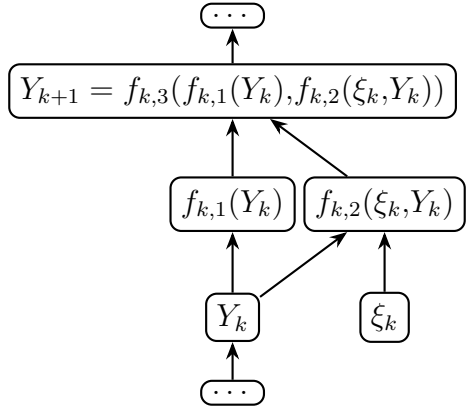
\begin{figure}[H]
    \centering
    \begin{tikzpicture}[rounded corners]
    \begin{scope}[every node/.style={rectangle,thick,draw}]
        \node[draw] (previous) at (0,0) {$Y_k$};
        \node[draw] (noise) at (2,0) {$\xi_k$};
        \node[draw] (next) at (0,3) {$Y_{k+1} = f_{k,3}(f_{k,1}(Y_k),f_{k,2}(\xi_k,Y_k))$};
        \node[draw] (unary) at (0,1.5) {$f_{k,1}(Y_k)$};
        \node[draw] (product) at (2,1.5) {$f_{k,2}(\xi_k,Y_k)$};
        \node[draw] (dots_south) at (0,-1) {\dots};
        \node[draw] (dots_north) at (0,4) {\dots};
    \end{scope}

    \begin{scope}[
                >={Stealth[black]},
                every node/.style={fill=white,rectangle},
                every edge/.style={draw,thick}
              ]
            \path [->] (previous) edge   (product); 
            \path [->] (previous) edge   (unary);
            \path [->] (unary)    edge (next);
            \path [->] (noise) edge  (product); 
            \path [->] (product) edge   (next); 
            \path [->] (dots_south) edge  (previous);
            \path [->] (next) edge  (dots_north);
    \end{scope}
    \end{tikzpicture}
  \caption{Section of the computational graph for Euler-Maruyama. Dots indicate the inductive continuation of the graph.}
  \label{f.euler-maruyama}
\end{figure}
The output of the computational graph is $Y_N$.

\subsubsection{Order statistics and bubble sort}
Let $n\geq 1$ and let $X_1,X_2,\dots,X_n$ be a sequence of random variables. The order statistics 
of $X_1,X_2,\dots,X_n$ is the random vector $(X_{(1)},X_{(2)},\dots,X_{(n)})$ defined as the result 
of applying a sorting algorithm to the random vector $(X_1,X_2,\dots,X_n)$. Hence, different 
sorting algorithms produce different DCGs. As a concrete example we consider the case of bubble 
sort described below.

Before proceeding, note that for any $i,j$ the random variables $X_{(i)}$ and $X_{(j)}$ are strongly dependent. Indeed, for any $i=1,2,\ldots,n$ the random variable $X_{(i)}$ is measurable with respect to $X_1,\dots,X_n.$

\begin{algorithm}[H]
\caption*{{\bf Bubble Sort}}
\begin{algorithmic}[1]
\State {\bf Input} : List $A$ of length $n$
\State {\bf Output}:  Sorted list $A$ in ascending order
\For{$i = 0$ to $n-1$ }
  \For{$j = 0$ to $n-i-2$}
    \If{$A(j) > A(j+1)$}
      \State Swap $A(j)$ and $A(j+1)$
    \EndIf
  \EndFor
\EndFor
\end{algorithmic}
\end{algorithm}
We shall rephrase this algorithm slightly as:
\begin{algorithm}[H]
\caption*{{\bf Bubble Sort} (Modified)}
\begin{algorithmic}[1]
\State {\bf Input} : List $A$ of length $n$
\State {\bf Output}:  Sorted list $A$ in ascending order
\For{$i = 0$ to $n-1$ }
  \For{$j = 0$ to $n-i-2$}
    \State $m \gets \min(A(j) ,A(j+1))$
    \State $M \gets \max(A(j) ,A(j+1))$
    \State $A(j) \gets m, A(j+1)\gets M$    
  \EndFor
\EndFor
\end{algorithmic}
\end{algorithm}
To visualize the computational graph of this algorithm we shall introduce some notation. For each $i=0,1,...,n-1$  define $(A(i,j))_{j=0}^{n-1}$ to be the partially sorted list after $i$ iterations of the outer loop and let $A(-1,j) = A(j)$. In practice, one does not construct this sequence of lists; they serve only as notational convenience.

Then for $i = 0,1,\dots,n-1$ and $j = 0,1,\dots,n-i-2$ we define
\begin{eqnarray}
        A(i,j) = \min(A(i-1,j),A(i-1,j+1)),\\
        A(i,j+1) = \max(A(i-1,j),A(i-1,j+1)),
\end{eqnarray}
while for $j=n-i,\dots,n-1$ we define
\begin{equation}
        A(i,j) = A(i-1,j),\quad j = n-i,\dots,n-1.
\end{equation}

\begin{figure}[H]
    \centering
    \begin{tikzpicture}[rounded corners]
    \begin{scope}[every node/.style={rectangle,thick,draw}]
        \node[draw] (dots_south) at (0,-1) {\dots};
        \node[draw] (previous_left) at (-3,0) {$A(i-1,j)$};
        \node[draw] (previous_right) at (3,0) {$A(i-1,j+1)$};
        \node[draw] (next_left) at (-3,4) {$A(i,j)$};
        \node[draw] (next_right) at (3,4) {$A(i,j+1)$};
        \node[draw] (min) at (-3,2) {$\min(A(i-1,j),A(i-1,j+1))$};
        \node[draw] (max) at (3,2) {$\max(A(i-1,j),A(i-1,j+1))$};
        \node[draw] (dots_north) at (0,5) {\dots};
    \end{scope}

    \begin{scope}[
                >={Stealth[black]},
                every node/.style={fill=white,rectangle},
                every edge/.style={draw,thick}
              ]
            \path [->] (previous_left) edge  (min); 
            \path [->] (previous_right) edge  (min); 
            \path [->] (previous_left) edge  (max); 
            \path [->] (previous_right) edge  (max); 
            \path [->] (min) edge (next_left);
            \path [->] (max) edge (next_right);
            \path [->] (dots_south) edge (previous_left);
            \path [->] (dots_south) edge (previous_right);
            \path [<-] (dots_north) edge (next_left);
            \path [<-] (dots_north) edge (next_right);
    \end{scope}
    \end{tikzpicture}
  \caption{Section of the computational graph for Bubble sort showing iteration $i$ for $j = 0,\dots,n-i-2$.}
\end{figure}
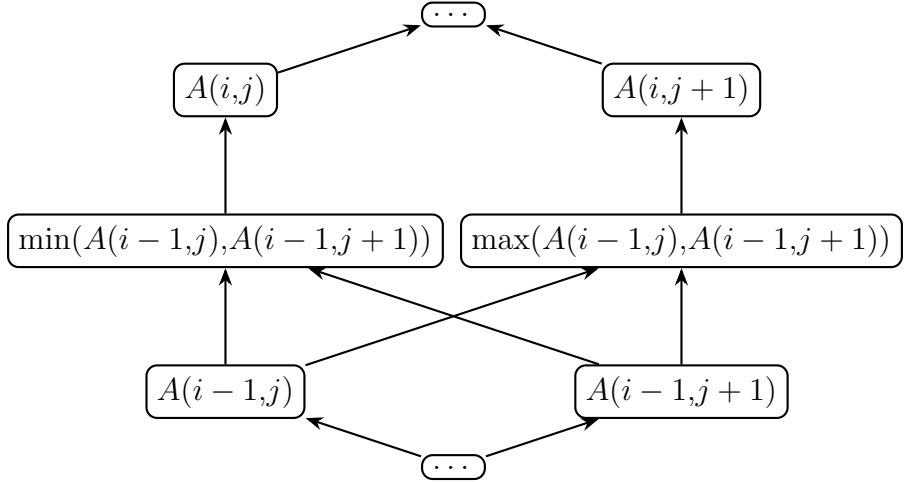
The output of the computational graph is the fully sorted list 
\[
A(n-1,0),A(n-1,1),\ldots,A(n-1,n-1),
\]
corresponding to the order statistics 
\[
(X_{(1)},X_{(2)},\dots,X_{(n)}).
\]

%% file: core.tex
\section{Upper Bound for General Computational Graphs}\label{s.upper-bound}
This section contains the first main result. We first consider the error induced between the DCG and the qDCG and then we compare the error between the qDCG and the cqDCG.
The idea of the proof is to use the recursive nature of the computational graph to obtain a bound on $W_1(\mu_\Delta,\mu_{\Delta}^{(n),c})$ in terms of $W_1(\mu_s,\mu_{s}^{(n),c}), s \in \SFS$. 
As both results follow similar ideas we formulate two general lemmas. We remark that Evans and Schulman~\cite[Lemma 2, p.2371]{796377} proved a similar result (but with a different proof) in the context of signal processing under noisy circuits. Evans et al.~\cite{MR1768240} also obtained a similar result (albeit in a more specialized setting).

For both proofs we shall use the following sets.
Define $C_\Delta^{(0)} = \{\Delta\},C_\Delta^{(1)}=\{\Delta\}\cup \inedges(\Delta)$ and recursively
\[
C_\Delta^{(n)}=C_\Delta^{(n-1)}\cup\bigcup_{v \in C_\Delta^{(n-1)}\setminus C_\Delta^{(n-2)}} \inedges(v).
\]
Note that for all $n\geq \depth(G)$ we have $C_\Delta^{(n)} = V$. We point out that if one considers the reversed DAG $\tilde{G} :=(V,\tilde{E})$ obtained by inverting all the edges 
\[
E \ni (u,v) \mapsto (v,u) \in \tilde{E},
\]
then the set $C_\Delta^{(n)}$ corresponds to the ball (defined by the directed metric) in $\tilde{G}$ centered at $\Delta$.
\begin{lemma}\label{l.output-bound}
    Suppose $G$ is a directed acyclic graph with the property that there exists a unique vertex 
    $\Delta \in V$ satisfying $\Delta \geq v, \forall v$. Let $c_v \geq 0, \forall v \in V$ and 
    assume that there exists $\rho_e \geq 0 ,\ e \in E$ such that 
    \begin{equation}\label{assumption.propagation-upper-bound}
        c_v \leq \sum_{u \in \inedges(v)} \rho_{(u,v)} c_u, \forall v \in V\setminus \SFS.
    \end{equation}
    Then 
    \begin{equation}
        c_\Delta \leq \sum_{s \in \SFS} \sum_{\gamma \in \SFP(s,\Delta)} \rho_\gamma c_s
    \end{equation}
    where $\rho_\gamma = \prod_{e \in \gamma}\rho_e.$
\end{lemma}
\begin{proof}
    Define 
    \[
    \Theta^{(n)}_\Delta = \sum_{v \in \inbdry C_\Delta^{(n)}} \sum_{\gamma \in \SFP(v,\Delta)}  \rho_\gamma c_v, n \geq 0.
    \]
    Note that by \ref{assumption.propagation-upper-bound} we have 
    \[
    \Theta_\Delta^{(n)} \leq \Theta_\Delta^{(n+1)}, \forall n \geq 0,
    \]
    and since $\Theta_\Delta^{(0)}= c_\Delta$ the result follows immediately as 
    \[
    c_\Delta = \Theta_\Delta^{(0)} \leq \Theta^{(\depth(G))} =  \sum_{s \in \SFS} \sum_{\gamma \in \SFP(s,\Delta)} \rho_\gamma c_s.
    \]
\end{proof}
We shall also use the following extension. While the general formulation may look cumbersome, when 
applied to our setting it simplifies considerably.
\begin{lemma}\label{l.output-bound-extended}
    Suppose $G$ is a directed acyclic graph with the property that there exists a unique vertex 
    $\Delta \in V$ satisfying $\Delta \geq v, \forall v$. 
    Let $c_v,d_v \geq 0, \forall v \in V$ and 
    assume that there exists $\rho_e,\phi_e \geq0 ,\ e \in E$ such that 
    \begin{equation}\label{assumption.propagation-upper-bound-coupled}
        \begin{split}            
        c_v &\leq \sum_{u \in \inedges(v)} \rho_{(u,v)} (c_u+d_u), \forall v \in V\setminus \SFS,\\
        d_v &\leq \sum_{u \in \inedges(v)} \phi_{(u,v)} d_u, \forall v \in V\setminus \SFS.
        \end{split}
    \end{equation}
    Then 
    \begin{equation}
        \begin{split}
            c_\Delta & \leq \sum_{s \in \SFS} \sum_{\gamma \in \SFP(s,\Delta)} \rho_\gamma c_s\\
                    &+
                    \sum_{s \in \SFS} 
                        d_s 
                        \sum_{n=0}^{\depth(G)-1} 
                            \sum_{u \in \inbdry C_\Delta^{(n+1)}}
                                \sum_{
                                    \substack{
                                        \gamma \in \SFP(s,u)\\ 
                                        \gamma' \in \SFP(u,\Delta)
                                        }
                                }
                                    \phi_\gamma \rho_{\gamma'}
        \end{split}
    \end{equation}
    where $\phi_\gamma = \prod_{e \in \gamma} \phi_e ,\ \rho_\gamma = \prod_{e \in \gamma'}\rho_v.$
\end{lemma}
\begin{proof}
    By Lemma \ref{l.output-bound} we have
    \begin{equation}
        d_v \leq \sum_{s \in \SFS} \sum_{\gamma \in \SFP(s,v)} \phi_\gamma d_s =: D_v, \forall v \in V \setminus \SFS.
    \end{equation}

    Now define 
    \begin{equation}
        \Xi^{(n+1)}_\Delta 
        = 
        \Xi^{(n)}_\Delta 
        + 
        \sum_{u \in \inbdry C_\Delta^{(n+1)}} 
            \sum_{\gamma \in \SFP(u,\Delta) } 
                \rho_\gamma D_u, \ \Xi^{(0)}= 0
    \end{equation}
    and 
    \begin{equation}
        \Theta^{(n)} 
        = 
        \sum_{v \in \inbdry C_\Delta^{(n)}} 
            \sum_{\gamma \in \SFP(v,\Delta) } 
                \rho_\gamma c_v
        + \Xi_\Delta^{(n)}, \Theta^{(0)} = c_\Delta.
    \end{equation}
    By Equation \eqref{assumption.propagation-upper-bound-coupled} it follows immediately that
    \begin{equation}
        \Theta^{(n)} \leq \Theta^{(n+1)},
    \end{equation}
    which implies that 
    \[
    c_\Delta \leq 
    \sum_{v \in \inbdry C_\Delta^{(n)}} 
            \sum_{\gamma \in \SFP(v,\Delta) } 
                \rho_\gamma c_v + \Xi_\Delta^{(\depth(G))}.
    \]
    Finally, by the definition of $\Xi^{(n)}_\Delta$ and $D_u$ we get 
    \begin{align*}
    \Xi_\Delta^{(\depth(G))} 
    &= 
    \sum_{n=0}^{\depth(G)-1} \Xi^{(n+1)}_\Delta -\Xi^{(n)}_\Delta 
    = 
    \sum_{n=0}^{\depth(G)-1} \sum_{u \in \inbdry C_\Delta^{(n+1)}} 
            \sum_{\gamma \in \SFP(u,\Delta) } 
                \rho_\gamma D_u\\
    &=    \sum_{n=0}^{\depth(G)-1} \sum_{u \in \inbdry C_\Delta^{(n+1)}} 
            \sum_{\gamma \in \SFP(u,\Delta) } 
                \rho_\gamma \sum_{s \in \SFS} \sum_{\gamma \in \SFP(s,u)} \phi_\gamma d_s \\
    \end{align*}
    and the result follows by re-arranging the sums.
\end{proof}

\subsection{Approximation errors}\label{ss:approximation_errors}
We begin with establishing the error between the DCG and the qDCG.
\begin{lemma}\label{l.quantized_recursive_error}
    Let $(G,\CF)$ be a computational graph with terminal vertex $\Delta$ and sources $\SFS$. Then 
    \begin{equation}\label{e.quantized_recursive_error}
        W_{1}(\mu_\Delta,\mu_\Delta^{(n)}) \leq \sum_{s \in \SFS} \sum_{\gamma \in \SFP(s,\Delta)} W_1(\mu_s,\mu_s^{(n)}) \prod_{e \in \gamma} \|f_{e_+}\|_\Lip .
    \end{equation}    
\end{lemma}
\begin{proof}
Observe that by the Lipschitz property we have     
\begin{align*}
    \BE[
        \left|
            X_v -X_v^{(n)} 
        \right|
        ] &= \BE[|f_v(\{X_u : u \in \inedges(v)\})-f_v(\{X_u^{(n)} : u \in \inedges(v)\}) |]\\
    &\leq \sum_{u\in \inedges(v)} \|f_v\|_\Lip \BE[|X_u-X_u^{(n)}|],
\end{align*}
from which it follows
\begin{equation}
\begin{split}
    W_1(\mu_v,\mu_v^{(n)})  \leq \sum_{u\in \inedges(v)} \|f_v\|_\Lip W_1(\mu_u,\mu_u^{(n)})  .
\end{split}
\end{equation}
Hence by Lemma \ref{l.output-bound} we deduce the result.
\end{proof}
\begin{remark}
    The argument above  shows that the output of a 
    distributional computational graph is Lipschitz continuous as a function of $\mu_s, s \in \SFS$ 
    with respect to the $W_1$.
\end{remark}

We now establish the analogous result between compressed-and-quantized DCG's.
\begin{lemma}\label{l.compressed_quantized_recursive_error}
Let $(G,\CF)$ be a computational graph with terminal vertex $\Delta$ and sources $\SFS$. Then 
\begin{equation}\label{e.compressed_quantized_recursive_erro}
W_1
\left(
    \mu_\Delta^{(n)},\mu_\Delta^{(n),c}
\right) \leq \frac{3}{2^{n+1}}
\sum_{s \in \SFS}
    \sum_{\gamma \in \SFP(s,\Delta)} 
    \diam(\supp(\mu_s^{(n)})) |\gamma|
        \prod_{e \in \gamma} 
            \|f_{e_+}\|_\Lip 
\end{equation}
\end{lemma}
\begin{proof}
First note that by the triangle inequality
\begin{align*}
    \BE\left[
            \left|
                X_v^{(n)} - X_v^{(n),c}
            \right| 
        \right] &= 
    \BE
        \left[
            \left|
                f_v\left(\{X_u^{(n)} : u \in \inedges(v) \}\right)
                - T \left(
                    f_v\left(\{X_u^{(n),c} : u \in \inedges(v) \}\right),
                    n
                    \right)
            \right| 
        \right] \\
    & \leq \BE\left[
            \left|
                f_v\left(\{X_u^{(n)} : u \in \inedges(v) \}\right)
                - 
                f_v\left(\{X_u^{(n),c} : u \in \inedges(v) \}\right)    
            \right| 
        \right] \\
    &\quad + 
    \BE\left[
            \left|
                f_v\left(\{X_u^{(n),c} : u \in \inedges(v) \}\right)
                - T \left(
                    f_v\left(\{X_u^{(n),c} : u \in \inedges(v) \}\right),
                    n
                    \right)
            \right| 
        \right].
\end{align*}
Now, we  upper bound the first term by
\begin{equation}\label{e:first_term_cq_bound}
\begin{split}
    &\BE\left[
            \left|
                f_v\left(\{X_u^{(n)} : u \in \inedges(v) \}\right)
                - 
                f_v\left(\{X_u^{(n),c} : u \in \inedges(v) \}\right)    
            \right| 
        \right] \\
    &\leq     \|f_v\|_\Lip \sum_{u\in \inedges(v)}
    \BE\left[
            \left|
                X_u^{(n)}
                - 
                X_u^{(n),c}    
            \right| 
        \right] .
\end{split}    
\end{equation}
For the second term we see that by Equation \ref{e.discrete_upper_bound} and the fact that $ \diam(\supp(\mu^{(n),c}_v)) \leq  \diam(\supp(\mu^{(n)}_v))$ we get 
\begin{equation}\label{e:second_term_cq_bound}
    \begin{split}    
        &\BE\left[
                \left|
                    f_v\left(\{X_u^{(n),c} : u \in \inedges(v) \}\right)
                    - T \left(
                        f_v\left(\{X_u^{(n),c} : u \in \inedges(v) \}\right),
                        n
                        \right)
                \right| 
            \right] \\ 
            &\overset{\eqref{e.discrete_upper_bound}}{\leq} 
            \frac{
                \diam(\supp(\mu^{(n),c}_v))
                }{2^{n+1}} \leq
            \frac{
                \diam(\supp(\mu^{(n)}_v))
                }{2^{n+1}} .
    \end{split}
\end{equation}
Now, since $X_v^{(n)} = f_v (\{ X_u^{(n)} : u \in \inedges(v)\})$ it follows that 
\begin{equation}\label{e:diam_supp_lipschitz_bound}
\diam(\supp(\mu^{(n)}_v)) \leq \|f_v\| \sum_{u \in \inedges(v)}\diam(\supp(\mu^{(n)}_u)).
\end{equation}
Hence if
\begin{equation}
    \begin{split}        
    c_v := 
    \frac{
            \diam(\supp(\mu^{(n)}_v))
    }{2^{n+1}} + 
    \BE\left[
            \left|
                X_v^{(n)}
                - 
                X_v^{(n),c}    
            \right| 
        \right],
    d_v :=   
    \frac{
            \diam(\supp(\mu^{(n)}_v))
    }{2^{n+1}}
    \end{split}
\end{equation}
then
\begin{equation}\label{e.crude-bound}
\begin{split}
    c_v &\overset{\eqref{e:first_term_cq_bound},\eqref{e:second_term_cq_bound}}{\leq} \|f_v\|_\Lip \sum_{u \in \inedges(v)}
    \BE\left[
            \left|
                X_u^{(n)}
                - 
                X_u^{(n),c}    
            \right| 
        \right] 
    +2\frac{
            \diam(\supp(\mu^{(n)}_v))
    }{2^{n+1}} \\
    &\overset{\eqref{e:diam_supp_lipschitz_bound}}{\leq}
    \|f_v\|_\Lip \sum_{u \in \inedges(v)}\BE\left[
            \left|
                X_u^{(n)}
                - 
                X_u^{(n),c}    
            \right| 
        \right]
        + 2 \|f_v\|_\Lip \sum_{u \in \inedges(v)}\frac{
            \diam(\supp(\mu^{(n)}_u))
    }{2^{n+1}} \\
        &= \|f_v\| \sum_{u \in \inedges(v)} c_u+d_u.
\end{split}
\end{equation}

Hence, by Lemma \ref{l.output-bound-extended} we can deduce that 
\begin{equation}\label{e.almost-last}
\begin{split}
&W_1
\left(
    \mu_\Delta^{(n)},\mu_\Delta^{(n),c}
\right) \\
&\leq 
\sum_{s \in \SFS} 
\left[
\BE
    \left[
        \left|
            X_s^{(n)}-X_s^{(n),c}
        \right|
    \right]
+\frac{\diam(\supp(\mu_s^{(n)}))}{2^{n+1}}\right]
    \sum_{\gamma \in \SFP(s,\Delta)} 
        \prod_{e \in \gamma} 
            \|f_{e_+}\|_\Lip \\
&+
\sum_{s \in \SFS} 
\frac{\diam(\supp(\mu_s^{(n)}))}{2^{n+1}}
    \sum_{m=0}^{\depth(G)-1} 
        \sum_{u \in \inbdry C_\Delta^{(m+1)}}  
            \sum_{
                \substack{
                    \gamma \in \SFP(s,u)\\ 
                    \gamma' \in \SFP(u,\Delta)
                    }
                }
                \prod_{e \in \gamma}   \|f_{e_+}\|_\Lip
                \prod_{e \in \gamma'}  \|f_{e_+}\|_\Lip.
\end{split}
\end{equation}
Now, we observe that the last sum is nothing but a partition
\begin{equation}\label{e.partition}
\sum_{m=0}^{\depth(G)-1} 
\sum_{u \in \inbdry C_\Delta^{(m+1)}}  
    \sum_{
        \substack{
            \gamma \in \SFP(s,u)\\ 
            \gamma' \in \SFP(u,\Delta)
            }
        }
        \prod_{e \in \gamma}   \|f_{e_+}\|_\Lip
        \prod_{e \in \gamma'}  \|f_{e_+}\|_\Lip
=
\sum_{\gamma \in \SFP(s,\Delta)} |\gamma| \prod_{e \in \gamma} \left\|f_{e_+}\right\|_{\Lip}.
\end{equation}
Thus, by finally observing that
\begin{align}\label{e.even-almost-laster}
\BE
\left[
    \left|
        X_s^{(n)}-X_s^{(n),c}
    \right|
\right]
+\frac{\diam(\supp(\mu_s^{(n)}))}{2^{n+1}}
\overset{\eqref{e.discrete_upper_bound}}{\leq} \frac{ \diam(\supp(\mu_s^{(n)}))}{2^n}
\end{align}
we can deduce the result
\begin{equation}
W_1
\left(
    \mu_\Delta^{(n)},\mu_\Delta^{(n),c}
\right) \overset{\eqref{e.almost-last},\eqref{e.even-almost-laster},\eqref{e.partition}}{\leq} \frac{3}{2^{n+1}}
\sum_{s \in \SFS}
    \sum_{\gamma \in \SFP(s,\Delta)} 
    \diam(\supp(\mu_s^{(n)})) |\gamma|
        \prod_{e \in \gamma} 
            \|f_{e_+}\|_\Lip .
\end{equation}
\end{proof}
Putting these two results we can now prove our first main result.
\begin{proof}[Proof of Theorem~\ref{thm.upper-bound}]
    The result follows easily from the triangle inequality 
    \begin{align*}
        &W_{1}(\mu_\Delta,\mu_\Delta^{(n),c})  \leq W_{1}(\mu_\Delta,\mu_\Delta^{(n)})+W_{1}(\mu_\Delta^{(n)},\mu_\Delta^{(n),c}) \\
        &\overset{\eqref{e.quantized_recursive_error},\eqref{e.compressed_quantized_recursive_erro}}{\leq} \sum_{s \in \SFS} \sum_{\gamma \in \SFP(s,\Delta)} W_1(\mu_s,\mu_s^{(n)}) \prod_{e \in \gamma} 
            \|f_{e_+}\|_\Lip \\
        & \quad+
        \frac{3}{2^{n+1}}\sum_{s \in \SFS}
        \sum_{\gamma \in \SFP(s,\Delta)} 
            \diam(\supp(\mu_s^{(n)})) 
                |\gamma|
                \prod_{e \in \gamma} 
                    \|f_{e_+}\|_\Lip  \\
        &\leq \sum_{s \in \SFS} \sum_{\gamma \in \SFP(s,\Delta)}
        \left(
            W_1(\mu_s,\mu_s^{(n)})+\frac{3|\gamma|}{2^{n+1}}\diam(\supp(\mu_s^{(n)})) 
        \right)
        \prod_{e \in \gamma} 
            \|f_{e_+}\|_\Lip.
    \end{align*}
\end{proof}
As a corollary we can deduce a certain optimality result (see Remark \ref{r.optimal-rate} just below this result for an explanation) for compactly supported distributions. 
\begin{corollary}
    Assume that $\supp(\mu_s)$ is compact for all $s \in \SFS$. Then for any computational graph $(G,\CF)$ there exists a constant $c = c(G,\CF,(\mu_s)_{s\in \SFS})>0$ such that 
    \begin{equation}
        W_1(\mu_\Delta,\mu^{(n),c}_\Delta) \leq \frac{c}{2^n}.
    \end{equation}
\end{corollary}
\begin{proof}
    The proof follows immediately from \cite[Theorem 4.2]{bilgin2025quantizationprobabilitydistributionsdivideandconquer} which provides the upper bound 
    \[
    W_1(\mu_s,\mu_s^{(n)}) \leq \frac{\diam(\supp(\mu_s))}{2^{n+1}}.
    \]
    Hence 
    \begin{align*}
        W_1(\mu_\Delta,\mu^{(n),c}_\Delta) 
        &\leq  \frac{1}{2^{n-1}} \max_{s \in \SFS} \diam(\supp(\mu_s)) \sum_{s \in \SFS} \sum_{\gamma \in \SFP(s,\Delta)} \prod_{e \in \gamma} \|f_{e_+}\|_\Lip  =: \frac{c}{2^n}.
    \end{align*}
\end{proof}
\begin{remark}\label{r.optimal-rate}
We now clarify what we mean by the optimality of the previous result.
A fundamental result in quantization theory is Zador's theorem.
\begin{theorem}[Zador's Theorem, {\cite[Theorem~6.2, p.~78]{MR1764176}}]\label{thm.zador}
    Let $X$ be a random variable with distribution $\mu$ that is absolutely continuous with respect to Lebesgue measure, having density $f$. If $\mathbb{E}[|X|^{1+\delta}] < \infty$ for some $\delta > 0$, then
    \begin{equation*}
        \lim_{n\to \infty} n \inf_{\substack{\nu \in \mathcal{M} \\ |\supp(\nu)| = n }} W_{1}(\mu,\nu) = \frac{1}{4} 
        \left(
            \int \sqrt{f(x)} \, dx 
        \right)^2,
    \end{equation*}
    where the infimum is over all discrete probability measures supported on exactly $n$ points.
\end{theorem}
Hence, there exists some $c(\mu)>0$ such that 
\[
W_{1}(\mu,\mu^{(n)}) \geq \frac{c(\mu)}{2^n}.
\]
Thus, the upper bound~\eqref{e.discrete_upper_bound} is optimal in terms of the rate of decay, though the constant is likely not sharp.
\end{remark}

\section{Error Bounds for the Euler-Maruyama Scheme}\label{s:em-bound}
We end this article by proving an extension of Theorem~\ref{thm.upper-bound} by applying the DCG framework to the Euler-Maruyama scheme. Since the function $f_{k,2}$ (see Equation \eqref{e.em-functions}) of the Euler-Maruyama computational graph, Figure \ref{f.euler-maruyama}, is not Lipschitz-continuous, Theorem~\ref{thm.upper-bound} is not directly applicable. 
To simplify the presentation we add the following assumption. Define $(X_t)_{t \geq 0}$  as the solution to the SDE 
\begin{equation*}
    \id X_t = a(t,X_t) \id t + b(t,X_t) \id W_t, t \in [0,T], X_0 = 0,
\end{equation*}
satisfying Assumption \ref{a.sde-assumptions}.
\begin{assumption}\label{a.sde-symmetry}
For all $x \in \BR$ and $t \in [0,T]$ we have
    \begin{equation}
    \begin{split}
        a(t,-x) &= -a(t,x),\\
        b(t,-x) &= b(t,x).
    \end{split}
\end{equation}    
\end{assumption}
A consequence of this assumption is that the Euler-Maruyama approximation $(Y_k)_{k=0}^N$ is statistically symmetric in the sense $\BP( Y_k \in \cdot) = \BP ( -Y_k \in \cdot), \forall k = 0,1,2,\dots, N$. This implies that the support of the quantized distributions $(Y_k^{(n)})_{k=0}^N$ are symmetric about the origin.
Additionally throughout this section we shall with out loss of generality assume that 
\begin{equation}
    \Delta t < 1.
\end{equation}

\subsection{Quantization error for Gaussian distribution}
Before we can turn to the main proof we need a result for the normal distribution related to the quantization error.
\begin{lemma}\label{l.gaussian-sequence}
    Let $X \sim \mathcal{N}(0,1)$ and define the sequence $(\omega_j)_{j\geq 0} $ by $\omega_0 = \BE[X], \omega_{j+1} = \BE[X | X \geq \omega_j]$. Then 
    \[
    \omega_n \sim \sqrt{2n}, \ n \to \infty.
    \]
\end{lemma}
\begin{proof}
    It should be evident from the definition of $\omega_j$ that the sequence is strictly increasing to $\infty$. For a proof of this fact we refer to \cite[Corollary 4.2]{bilgin2025quantizationprobabilitydistributionsdivideandconquer}. 
    To proceed we shall estimate the difference 
    \[
    \omega_{j+1}-\omega_j = \BE[X-\omega_j | X \geq \omega_j].
    \]
    as $j \to \infty.$
    Thus, we begin with analyzing the expression
    \begin{equation}\label{e.conditioned-gauss-moment}
        \BE[X -x | X \geq x ] = \frac{\varphi(x)}{\bar{\Phi}(x)} -x.
    \end{equation}
    For this, we use the asymptotic expansion for the tail $\bar{\Phi}(x) := \BP(X>x)$, \cite[Equation 7.12.1]{NIST:DLMF}. Let $\varphi(x) = \e^{-x^2/2}/\sqrt{2 \pi}$ denote the pdf of a standard normal random variable then for any $n \geq 1$
    \begin{equation}\label{e.asymptotic-normal-expression}
        \bar{\Phi}(x)  = \varphi(x) P_n(1/x)+O(\varphi(x) x^{-(2n+1)}),\ x \to \infty,
    \end{equation}
    where $P_n(x) = \sum_{k=0}^{n-1}(-1)^k (2k-1)!! x^{2k+1}$ and $ n!! = \prod_{k=1}^{\frac{n+1}{2}}(2k-1)$.
    By choosing $n = 3$ in \eqref{e.asymptotic-normal-expression} we get 
    \begin{align}\label{e.asymp-order}
    \nonumber
    \bar{\Phi}(x) &= \varphi(x) \left(\frac{1}{x}-\frac{1}{x^3}\right)( 1+\Theta(x^{-4}))+O(\varphi(x)x^{-7})\\
    &= \varphi(x) \left(\frac{1}{x}-\frac{1}{x^3}\right)( 1+O(x^{-4})).
    \end{align}
    Thus
    \begin{align}        
        \BE[X -x | X \geq x ] & 
        \overset{\eqref{e.conditioned-gauss-moment}}{=}
        \frac{\varphi(x)}{\bar{\Phi}(x)} -x  
        \overset{\eqref{e.asymp-order}}{=} 
        \frac{x^4 - x (x^3-x) ( 1-O(x^{-4}))}{(x^3-x) ( 1-O(x^{-4}))}\\
        &= \frac{x^2+O(1)}{x^3-x} \sim \frac{1}{x},\ x \to \infty.
    \end{align}
    Hence for any $\epsilon >0$ there exists an $N \geq 1$ such that 
    \begin{equation}
        \frac{1-\epsilon}{\omega_{j}} \leq \omega_{j+1}-\omega_{j} \leq \frac{1+\epsilon}{\omega_{j}}, \forall j \geq N.
    \end{equation}
    Now we observe that 
    \begin{align}
         \omega_j^2 + 2(1-\epsilon) + \frac{(1-\epsilon)^2}{\omega_j^2} \leq \omega_{j+1}^2 \leq \omega_j^2 + 2(1+\epsilon) + \frac{(1+\epsilon)^2}{\omega_j^2}, \forall j \geq N.
    \end{align}
    Since $\omega_j \to \infty, j \to \infty$ we can choose $N$ so large that $\frac{1}{\omega_j^2} \leq \frac{2\epsilon}{(1+\epsilon)^2}$. Since $\frac{(1-\epsilon)^2}{\omega_j^2}\geq 0$ and the aforementioned fact it follows that 
    \begin{align}
         \omega_j^2 + 2(1-\epsilon)  \leq \omega_{j+1}^2 \leq \omega_j^2 + 2(1+2\epsilon), \forall j \geq N.
    \end{align}
    Hence the squared increments $\omega_{j+1}^2-\omega_{j}^2$ are almost constant. Thus by writing $\omega_{j+1}^2$ as a telescoping sum
    \begin{equation}        
    \omega_{j+1}^2 = \sum_{i=N+1}^{j+1} \omega_i^2-\omega_{i-1}^2 + \omega_N^2 
    \end{equation}
    we deduce  
    \begin{equation}        
    2(1- \epsilon) (j-N+1) +\omega_N^2 \leq \omega_{j+1}^2 \leq 2(1+2 \epsilon) (j-N+1) +\omega_N^2.
    \end{equation}
    Thus, 
    \begin{equation}
        \begin{split}
            &\liminf_{j \to \infty} \frac{\omega_{j}^2}{j} \geq 2(1-\epsilon), \\
            &\limsup_{j \to \infty} \frac{\omega_{j}^2}{j} \leq 2(1+2\epsilon), 
        \end{split}
    \end{equation}
    and since $\epsilon>0$ was arbitrary the result follows. 
\end{proof}
Using this lemma, we can apply 
\cite[Theorem~1.1]{bilgin2025quantizationprobabilitydistributionsdivideandconquer}
to conclude that the Wasserstein-1 distance between $\mu(\id x) = \varphi(x) \id x$ and $\mu^{(n)}$ 
achieves optimal rate of decay (as discussed in Remark~\ref{r.optimal-rate}). 
\begin{corollary}\label{c.gauss-rate}
    Let $\varphi(x) =\frac{1}{\sqrt{2\pi}} \e^{-\frac{x^2}{2}} $ and  $\mu(\id x) =\varphi(x) \id x $ denote the law of a standard Gaussian random variable. Then
    \begin{equation}\label{e.gauss-rate}
    W_1(\mu,\mu^{(n)}) = O(2^{- n}).
    \end{equation}
\end{corollary}
\begin{proof}
Since $\mu$ is symmetric it follows that
\[    
W_1(\mu,\mu^{(n)}) =  W_1(\mu_+,\mu_+^{(n-1)}).
\]
Hence, the result follows once we establish an error bound between $\mu_+(\cdot) := \mu( \cdot |\BR_+)$ and $\mu_+^{(m)}$ for $m \geq 1$. 
Let $X_+ \sim \mu_+$ and define $\omega_{-1}= 0,\ \omega_0 = \int_{\BR_+} t \id \mu_+(t)=1$ and define $\omega_{j+1} = \BE[X_+ | X_+ \geq \omega_{j}], j\geq 0$ and $\Omega_j :=[\omega_{j-1},\omega_{j}]$. Then by \cite[Theorem 1.1]{bilgin2025quantizationprobabilitydistributionsdivideandconquer} the Wasserstein distance between $\mu_+$ and $\mu_+^{(m)}$ is 
\begin{equation*}
    \begin{split}        
        W_1(\mu_+,\mu_+^{(m)}) 
        \leq
        \sum_{j=0}^{m-1} 
            \frac{
                    (\omega_{j}-\omega_{j-1})\mu_+(\Omega_{j})
                }
                {2^{m-j}} 
        + \BE[ |X_+ - \omega_{m}| ; X_+ \geq \omega_{m-1}].
    \end{split}
\end{equation*}
Now by the standard Gaussian tail bound
\[
\BP(X \geq x) \leq \frac{\varphi(x)}{x}, x \geq 0,
\]
we have for $j \geq 1, m \geq 1$
\begin{equation}\label{e.gaussian-upper-bounds-1}
        \mu_+(\Omega_j) 
            \leq \mu_+[\omega_{j-1},\infty) = 2  \mu[\omega_{j-1},\infty) \leq 2 \frac{\varphi(\omega_{j-1})}{\omega_{j-1}},
\end{equation}
and for $m\geq 1$ we similarly have
\begin{equation}\label{e.gaussian-upper-bounds-2}
    \begin{split}
        &\BE[ |X_+ - \omega_{m}| ; X_+ \geq \omega_{m-1}] = 
        2 \BE[ |X - \omega_{m}| ; X \geq \omega_{m-1}]  \\
        & \leq 2 \BE[ X - \omega_{m-1} ; X \geq \omega_{m-1}] 
        \overset{\eqref{e.conditioned-gauss-moment}}{=}
        2 (\varphi(\omega_{m-1})-\omega_{m-1} \bar{\Phi}(\omega_{m-1}))
        \leq 2 \varphi(\omega_{m-1})
    \end{split}
\end{equation}
Hence,
\begin{equation}
    \begin{split}
         W_1(\mu_+,\mu_+^{(m)}) 
        \overset{\eqref{e.gaussian-upper-bounds-1},\eqref{e.gaussian-upper-bounds-2}}{\leq} 2 \sum_{j=0}^{m-1} \frac{ (\omega_j/(\omega_{j-1}\vee 1))\varphi(\omega_j)}{2^{m-j}} +2 \varphi(\omega_{m-1}).
    \end{split}
\end{equation}
Note that the $\omega_{j-1}\vee 1$ is just cosmetic and only affects the case $j=0$, where we in fact have $1= \omega_0-\omega_{-1} = \omega_0/(\omega_{-1}\vee 1).$

By Lemma \ref{l.gaussian-sequence} we see that for each $\epsilon > 0 $  there exists $N \geq 1$ such that  
\[
\sqrt{2(1-\epsilon)j} \leq \omega_j \leq \sqrt{2(1+\epsilon)j}, \forall j \geq N.
\]
Hence if we take $\epsilon>0$ so small that $\e^{-(1-\epsilon)} < \frac{1}{2}$ we see that for $m$ large enough there exist $c,c',c''>0$ dependent on $N,\epsilon$ such that
\begin{equation}
\begin{split}
    &\sum_{j=0}^{m-1} \frac{ (\omega_j/(\omega_{j-1}\vee 1))\varphi(\omega_j)}{2^{m-j}}
    \leq 
    \sum_{j=0}^{N} \frac{c}{2^{m-j}}  
    +
    \sum_{j=N+1}^{m-1} \frac{c' \e^{-(1-\epsilon)j} }{2^{m-j}} \leq c'' 2^{-m}.
\end{split}
\end{equation}
Finally, we observe that since
\begin{equation}
    2 \varphi(\omega_{m-1}) \leq c \e^{-m(1-\epsilon)}
\end{equation}
we can conclude
\begin{equation}
    W_1(\mu_+,\mu_+^{(m)}) = O(2^{- m}),
\end{equation}
which concludes the proof.
\end{proof}
\subsection{Error bound for the Euler-Maruyama scheme}
We divide the argument into two lemmas, similar to the approach in 
Section~\ref{ss:approximation_errors}.
\begin{lemma}\label{l.em-quant}
    Let $Y_j, j = 1,2,\dots,N$ denote the Euler-Maruyama scheme \eqref{def.euler-maruyama} and let $Y_j^{(n)},j=1,2,\dots,N$ denote the quantized version. Then there exists a constant $c = c( T, K ) > 0$ such that
    \begin{equation}
        \BE
            \left[
                \left|
                    Y_k-Y_{k}^{(n)}
                \right|
            \right] 
            \leq 
            \left(
                1+c \sqrt{\Delta t}
            \right)^k 
            \BE
                \left[
                    \left|
                        \xi_0-\xi_0^{(n)}
                    \right|
                \right],
    \end{equation}
    for all $k = 0,1,\dots,N.$
\end{lemma}
\begin{proof}
    Recall the definition of $f_{k,i}, i= 1,2,3$ from \eqref{e.em-functions} and for brevity let 
    \[
        Y_{k,1} = f_{k,1}(Y_k),Y_{k,2} = f_{k,2}(Y_k,\xi_k),
    \]
    so that $Y_{k+1} = f_{k,3}(Y_{k,1},Y_{k,2})$. For the first term we get 
\begin{align}\label{e.em-quantized-first-bound}
\nonumber
&\BE
\left[
    \left| 
        Y_{k+1}-Y_{k+1}^{(n)}
    \right|
\right] =  
\BE
\left[
    \left| 
         f_{k,3}
            \left(
                Y_{k,1},Y_{k,2}
            \right)
            - 
        f_{k,3}
            \left(
                Y_{k,1}^{(n)},Y_{k,2}^{(n)}
            \right)
    \right|
\right]
\\
\nonumber
&\leq
\BE
\left[
    \left| 
        Y_{k,1}-Y_{k,1}^{(n)}
    \right|
\right]+
\BE
\left[
    \left| 
        Y_{k,2}-Y_{k,2}^{(n)}
    \right|
\right] \\
&\leq (1+\Delta t K) 
\BE
\left[
    \left|
     Y_k-Y_k^{(n)}
    \right|
\right]
+\sqrt{\Delta t}
\BE
\left[
    \left|
        b(t_k,Y_k)\xi_k -b(t_k,Y_k^{(n)}) \xi_k^{(n)}
    \right|
\right].
\end{align}
So far we have followed the argument of Lemma \ref{l.quantized_recursive_error} but to estimate the 
last term of \eqref{e.em-quantized-first-bound} we utilize the linear growth assumption on $b$, see 
Assumption \ref{a.sde-assumptions}, as well as the independence between the $\xi_k$'s and $Y_k$'s.
\begin{align}\label{e.em-second-term-comp}
\nonumber
&\BE
\left[
    \left|
        b(t_k,Y_k)\xi_k -b(t_k,Y_k^{(n)}) \xi_k^{(n)}
    \right|
\right] \\
\nonumber
&\leq 
\BE
\left[
    \left|
        \xi_k^{(n)} 
        \left(
            b(t_k,Y_k) -b(t_k,Y_k^{(n)}) 
        \right)
    \right|
\right]
+
\BE
\left[
    \left|
        b(t_k,Y_k)(\xi_k - \xi_k^{(n)})
    \right|
\right]\\
&\leq K  \BE[|\xi_k^{(n)}|] \BE[|Y_k-Y_k^{(n)}|]+(1+K') \BE[|Y_k|] \BE[|\xi_k-\xi_k^{(n)}|].
\end{align}
Since $\sup_{t \in [0,T]} \BE[|X_t|] < \infty$ and by weak convergence
\[
|\BE[|Y_k|] - \BE[|X_{kT/N}|]| \leq \BE[|Y_k-X_{kT/N}|] \to 0 , N \to \infty, \forall k,
\]
we deduce that there exists $c = c(T)> 0$ such that $\BE[|Y_k|] \leq c , \forall k = 1,2,\ldots,N,\ \forall N \geq 1.$ Similarly, since $\xi_k^{(n)}$ converges weakly to $\xi_k$ we can bound $\BE[|\xi^{(n)}|] \leq c'$ by some universal constant $c'$. Hence 
\begin{align}
\BE
\left[
    \left| 
        Y_{k+1}-Y_{k+1}^{(n)}
    \right|
\right]
\leq 
\left(
    1+\Delta t K +\sqrt{\Delta t}K c'
\right)
\BE
\left[
    \left|
        Y_{k}-Y_{k}^{(n)} 
    \right|
\right]
+ c \sqrt{\Delta t}  
\BE
    \left[
        \left|
            \xi_k-\xi_k^{(n)}
        \right|
    \right].
\end{align}
Letting $a_k = \BE
\left[
    \left|
        Y_{k}-Y_{k}^{(n)} 
    \right|
\right],$ 
we see that the above is equivalent (since $\BE\left[ \left| \xi_k-\xi_k^{(n)}\right|\right]$ is independent of $k$) to 
\[
a_{k+1} \leq (1+\alpha) a_{k} +\beta, a_0 = \BE[|\xi_0-\xi^{(n)}_0|],
\]
with $\alpha = (1+ c  \sqrt{\Delta t}),\ \beta = c' \sqrt{\Delta t} \BE[|\xi_0-\xi_0^{(n)}|]$, where $c = c(K),\ c'= c'(K,T)$. Hence there exists some constant $c = c(T,K) >0$ such that 
\[
\BE[|Y_k-Y_{k}^{(n)}|] \leq \left(1+c \sqrt{\Delta t}\right)^k \BE[|\xi_0-\xi_0^{(n)}|].
\]
\end{proof}
Now we turn to the second lemma.
\begin{lemma}\label{l.em-quant-compr}
    Consider the quantized and the compressed-and-quantized version of the Euler-Maruyama 
    scheme. 
    Then there exists an absolute constant $c>0$ such that
    \begin{equation}
        \BE
        \left[ 
            \left| 
                Y_{k}^{(n)}-Y_{k}^{(n),c}
            \right|
        \right] \leq c \frac{( 1 +c \sqrt{n \Delta t})^{k+1} \sqrt{n}}{2^n}.
    \end{equation}
\end{lemma}
\begin{proof}
Recall once again the definition of $f_{k,i}, i= 1,2,3$ from \eqref{e.em-functions} and for brevity let 
\[
Y_{k,1} = f_{k,1}(Y_k), \ Y_{k,2} = f_{k,2}(Y_k,\xi_k),
\]
so that $Y_{k+1} = f_{k,3}(Y_{k,1},Y_{k,2}) = Y_{k,1}+Y_{k,2}$. Moreover, let $\mu_{k,i}$ denote the distribution of $Y_{k,i},$ along with the usual modifications $\mu_{k,i}^{(n)},\mu_{k,i}^{(n),c}.$
The proof follows closely that of Lemma \ref{l.compressed_quantized_recursive_error} but with some differences. By the triangle inequality we get
\begin{align*}
&\BE
\left[ 
    \left| 
        Y_{k+1}^{(n)}-Y_{k+1}^{(n),c}
    \right|
\right]\\
&\leq
\BE
\left[ 
    \left| 
        f_{k,3}(Y_{k,1}^{(n)},Y_{k,2}^{(n)})-f_{k,3}(Y_{k,1}^{(n),c},Y_{k,2}^{(n),c}) 
    \right|
\right]\\
&+        
\BE
\left[ 
    \left| 
        f_{k,3}(Y_{k,1}^{(n),c},Y_{k,2}^{(n),c}) - T(f_{k,3}(Y_{k,1}^{(n),c},Y_{k,2}^{(n),c}),n)
    \right|
\right]
\end{align*}
Now we observe that by Definition \ref{def.distributional-computational-graph} we have
\begin{equation}
Y_{k,1}^{(n)} = Y_{k,1}^{(n),c} = f_{k,1}(Y_k^{(n)}),
\end{equation}
which implies that we can bound the respective terms by
\begin{equation}
    \begin{split}
    &\BE
    \left[ 
        \left| 
            f_{k,3}(Y_{k,1}^{(n)},Y_{k,2}^{(n)})-f_{k,3}(Y_{k,1}^{(n),c},Y_{k,2}^{(n),c}) 
        \right|
    \right]
    \\
    &= 
    \BE
    \left[ 
        \left| 
            f_{k,2}(Y_{k}^{(n)},\xi_{k,2}^{(n)})-T(f_{k,2}(Y_{k}^{(n)},\xi_{k,2}^{(n)}),n)
        \right|
    \right]\\
    & \overset{\eqref{e.discrete_upper_bound}}{\leq} \frac{\diam(\supp(\mu_{k,2}^{(n)}))}{2^n}
    ,
    \end{split}
\end{equation}
and
\begin{equation}
    \begin{split}
    &\BE
    \left[ 
        \left| 
            f_{k,3}(Y_{k,1}^{(n),c},Y_{k,2}^{(n),c}) - T(f_{k,3}(Y_{k,1}^{(n),c},Y_{k,2}^{(n),c}),n)
        \right|
    \right]\\
    &=
    \BE
    \left[ 
        \left| 
            f_{k,3}(Y_{k,1}^{(n)},Y_{k,2}^{(n),c}) - T(f_{k,3}(Y_{k,1}^{(n)},Y_{k,2}^{(n),c}),n)
        \right|
    \right]\\
    & \overset{\eqref{e.discrete_upper_bound}}{\leq}\frac{\diam(\supp(\mu_{k,3}^{(n),c}))}{2^n} \leq  \frac{\diam(\supp(\mu_{k,3}^{(n)}))}{2^n}  .
    \end{split}
\end{equation}
We now begin estimating $\diam(\supp(\mu_{k,2}^{(n)}))$ .
Let $\mathcal{Y}_k^{(n)}$ denote the support of $Y_k^{(n)}$ and let $\mathcal{E}^{(n)}$ denote the support of $\xi_k^{(n)}$. Note that $\mathcal{E}^{(n)}$ is independent of $k$ and symmetric in the sense $\mathcal{E}^{(n)} = -\mathcal{E}^{(n)}$. Observe additionally that the symmetry also holds for $\mathcal{Y}_k^{(n)}$ in light of Assumption \ref{a.sde-symmetry}. Hence 
\begin{equation}\label{e.diam-k-2}
    \begin{split}
        &\diam(\supp(\mu_{k,2}^{(n)})) 
        = 
        \sup_{
            \substack{
                y,y' \in \mathcal{Y}_k^{(n)} \\ x,x' \in \mathcal{E}^{(n)} 
            }  
        } 
        | 
            f_{k,2}(y,x)-f_{k,2}(y',x')
        | 
        \\
        &= 
        \sqrt{\Delta t} 
        \sup_{
            \substack{
                y,y' \in \mathcal{Y}_k^{(n)} \\ x,x' \in \mathcal{E}^{(n)} 
            }  
        }
        |
            b(t_k,y)x-b(t_k,y')x'
        | \\
        &=
        \sqrt{\Delta t} 
        \sup_{
            \substack{
                y,y' \in \mathcal{Y}_k^{(n)} \\ x,x' \in \mathcal{E}^{(n)} 
            }  
        }
        |b(t_k,y)-b(t_k,y') | |x| + |x-x'||b(t_k,y')|
        \\
        &\leq 
        \sqrt{\Delta t} \diam(\mathcal{E}^{(n)})
        \sup_{
            \substack{
                y,y' \in \mathcal{Y}_k^{(n)} \\ x,x' \in \mathcal{E}^{(n)} 
            }  
        }
        |b(t_k,y)-b(t_k,y') | + K'(1+|y'|)\\
        & \leq c \sqrt{\Delta t} \diam(\mathcal{E}^{(n)}) \diam(\mathcal{Y}_k^{(n)})
    \end{split}
\end{equation}
In a similar vein we now turn to estimating the diameter of $\mathcal{Y}_{k+1}^{(n)} = \supp(\mu_{k,3}^{(n)})$.
\begin{equation}
    \begin{split}
        \diam(\mathcal{Y}_{k+1}^{(n)} )
        &= 
        \sup_{
            \substack{
                y,y' \in \mathcal{Y}_k^{(n)} \\ x,x' \in \mathcal{E}^{(n)} 
            }  
        }
        | 
            f_{k,3}(y,x) -f_{k,3}(y',x') 
        |
        \\
        & \leq
        \sup_{
            \substack{
                y,y' \in \mathcal{Y}_k^{(n)} \\ x,x' \in \mathcal{E}^{(n)} 
            }  
        }
        | 
            f_{k,1}(y) -f_{k,1}(y')
        |
        +
        | 
            f_{k,2}(y,x) -f_{k,2}(y',x') 
        |
        \\
        &\overset{\eqref{e.diam-k-2}}{\leq} (1+K\sqrt{\Delta t}) \diam(\mathcal{Y}_k^{(n)})+c \sqrt{\Delta t}\diam(\mathcal{E}^{(n)}) \diam(\mathcal{Y}_k^{(n)})
    \end{split}
\end{equation}
Now we note that by Lemma \ref{l.gaussian-sequence} there exists some absolute constant $c>0$ such that
\[
\diam(\mathcal{E}^{(n)}) \leq c \sqrt{n}, \forall n \geq 0,
\]
which yields
\begin{equation}
    \diam(\mathcal{Y}_{k+1}^{(n)} ) \leq (1+ c \sqrt{n \Delta t})  \diam(\mathcal{Y}_{k}^{(n)} ).
\end{equation}
Hence,
\begin{equation}
    \begin{split}
    \diam(\mathcal{Y}_{k}^{(n)} ) 
    &\leq (1+ c \sqrt{n \Delta t})^k  \diam(\mathcal{Y}_0^{(n)} ) 
    = (1+ c \sqrt{n \Delta t})^k  \diam(\mathcal{E}^{(n)} ) \\
    &\leq c (1+ c \sqrt{n \Delta t})^k \sqrt{n},
    \end{split}
\end{equation}
from which we can conclude
\begin{equation}
    \begin{split}
        \BE
        \left[ 
            \left| 
                Y_{k+1}^{(n)}-Y_{k+1}^{(n),c}
            \right|
        \right]
        \leq c \frac{( 1 +c \sqrt{n \Delta t})^{k+1} \sqrt{n}}{2^n}.
    \end{split}
\end{equation}
\end{proof}
We can now establish the main result.
\begin{proof}[Proof of Theorem \ref{thm.em-bound}]
Note that by the triangle inequality and Lemmas \ref{l.em-quant},\ref{l.em-quant-compr} we get the upper bound
\begin{equation}
    \begin{split}
        &\BE
        \left[
                \left|
                    Y_k-Y_{k}^{(n),c}
                \right|
        \right]
        \leq
        \BE
            \left[
                \left|
                    Y_k-Y_{k}^{(n)}
                \right|
            \right]
            +
        \BE
        \left[ 
            \left| 
                Y_{k}^{(n)}-Y_{k}^{(n),c}
            \right|
        \right]\\
        & \leq  \left(1+c \sqrt{\Delta t}\right)^k \BE[|\xi_0-\xi_0^{(n)}|]+c' \frac{( 1 +c' \sqrt{n \Delta t})^{k} \sqrt{n}}{2^n}.
    \end{split}
\end{equation}
Thus if we couple $\xi_0$ and $\xi_0^{(n)}$ using the optimal coupling provided by the Wasserstein distance we can apply
Corollary \ref{c.gauss-rate} and more specifically \eqref{e.gauss-rate} to deduce that there exists a constant $c$ such that 
\[
\BE[|\xi_0-\xi_0^{(n)}|] \leq \frac{c}{2^n}.
\]
Hence 
\begin{equation}
    \begin{split}
        &\BE
        \left[
                \left|
                    Y_k-Y_{k}^{(n),c}
                \right|
        \right] 
        \leq 
        \frac
            {
            \left(1+c \sqrt{\Delta t}\right)^k +c' ( 1 +c' \sqrt{n \Delta t})^{k} \sqrt{n}
            }{2^n}
            \\
        & \leq
        \frac
            {
                c ( 1 +c \sqrt{n \Delta t})^{k} \sqrt{n} 
            }{2^n} 
        \leq 
        \frac{c \e^{c ' k\sqrt{n \Delta t} }}{2^n}
    \end{split}
\end{equation}
where we used the elementary inequality 
\[
1+x \leq \e^x
\]
in the last inequality.
\end{proof}
\begin{remark}\label{r:em-bound}
    The $\sqrt{n}$ appearing in the exponent arises from the estimate $\diam(\mathcal{E}^{(n)}) \leq c \sqrt{n}$, which in turn is a consequence of Equation~\eqref{e.discrete_upper_bound}. We do not have numerical evidence to determine whether this dependence is tight.
    
    On the other hand, the term $k \sqrt{\Delta t}$ appears to be sharp, as indicated by the following numerical investigation.
    We consider the Euler-Maruyama approximation of a Geometric Brownian motion with 
    $\mu=0.05,\ \sigma = 0.4, \ Y_0 = 100, \ T =1$, and we plot the Wasserstein-1 distance between $Y_N$ and $Y_N^{(n),c}$ for $N=1,100,200,\ldots,1500$ with fixed $n$.
    \begin{figure}[H]
    \centering
    \begin{subfigure}{0.49\textwidth}
        \centering
        \includegraphics[width=\linewidth]{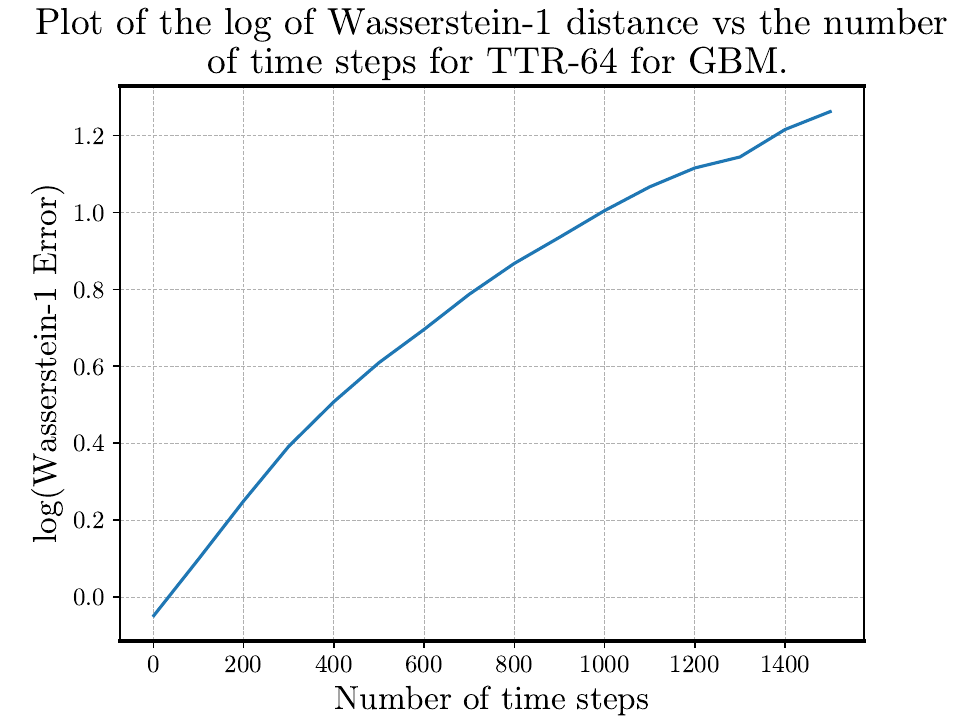}
    \end{subfigure} 
    \begin{subfigure}{0.49\textwidth}
        \centering
        \includegraphics[width=\linewidth]{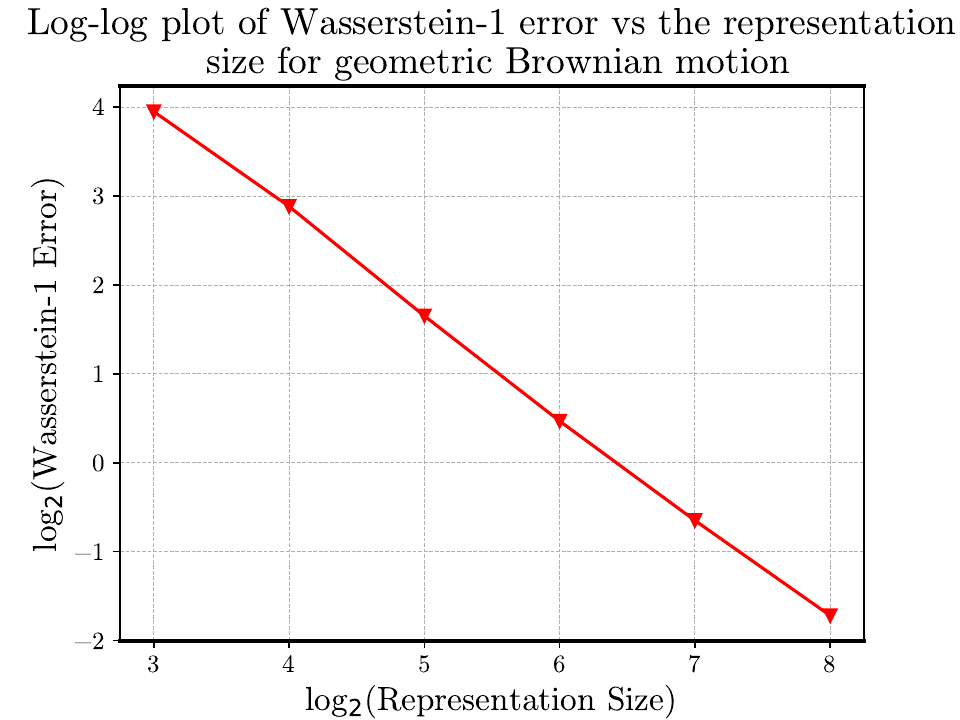}
    \end{subfigure}
    \caption{$\log W_1(Y_N, Y_N^{(n),c})$ as a function of $N$ (left) and $n$ (right).} 
    \label{fig:em-error}
    \end{figure}
    These figures suggest the bound is qualitatively tight. Indeed, the left figure shows that 
    $\log W_1(Y_N, Y_N^{(n),c})$ grows as $c N \sqrt{\Delta t} = c\sqrt{N}$ which is consistent 
    with the upper bound.
    The right figure demonstrates the optimal decay with respect to $n$ in the sense of 
    Theorem~\ref{thm.zador}.
\end{remark}

%% file: conclusions.tex
\section{Discussion}
\label{section:conclusions}
We first address the assumptions on the computational graph. The uniqueness assumption (that there 
exists a unique terminal vertex $\Delta \in V$) is purely cosmetic and simplifies the analysis. 
The Lipschitz assumption is more significant. Without it, one cannot guarantee that intermediate or 
terminal vertices have finite mean, which invalidates the entire argument. Stronger assumptions on 
the input distributions (such as finite moment generating functions) would allow for a larger class 
of functions.
We now examine the compression error from Remark \ref{rmk:compression} and discuss a possible 
improvement.

As observed in Remark \ref{r:cg-representation}, a natural additional assumption on the computational graph $(G,\CF)$ is
\begin{equation}
	\# \inedges(v) \leq 2, \forall v \in V.
\end{equation}
This implies that all functions $f \in \CF$ are bivariate. Hence, if the support size of the 
quantized input distributions is
\[
\# \supp(\mu_s^{(n)}) = 2^n, \ s \in \SFS,
\]
then for $\inedges(w) =\{u,v\}$ we have
\[
\# \supp(\mu_w^{(n),c}) \leq  \# \supp(\mu_u^{(n),c}) \cdot \# \supp(\mu_v^{(n),c}) \leq 2^{2n}.
\]
Thus as a first step in understanding the compression error, one could study how the error
\[
W_1(\mu^{(2n)},T(\mu^{(2n)},n))
\]
behaves as $n \to \infty$ for some absolutely continuous $\mu \in \CP_1$. By the triangle inequality, this yields
\[
W_1(\mu^{(2n)},T(\mu^{(2n)},n)) \leq W_1(\mu^{(2n)},\mu) + W_1(\mu,T(\mu^{(2n)},n)).
\]
The first term is the standard quantization error, so analyzing $W_1(\mu,T(\mu^{(2n)},n))$ may be 
more tractable.

As a broader observation, the theoretical results (Theorems~\ref{thm.upper-bound} and~\ref{thm.em-bound}) and Figure~\ref{fig:em-error} suggest that increasing $N$ 
does not necessarily improve the approximation. This points to a fundamental trade-off between 
approximation fidelity and the number of operations---or equivalently, the size 
of the computational graph.

We conclude by noting that the DCG framework depends on choosing a distributional representation 
and a compression algorithm. In this article we have studied a concrete compression algorithm, but 
depending on the problem structure, alternative representations and compression schemes may be more 
appropriate. The computational implications of these choices are addressed in companion work.

\subsection{Future work}
In a companion article we will focus more on the computational complexity and benchmarking against 
Monte Carlo programs.

Efficient higher-dimensional quantization algorithms suitable for compression remain 
an open research direction. Current methods scale poorly with dimension, and developing algorithms 
that balance computational cost with approximation accuracy is crucial for practical applications.

Extending the analysis to distributions with higher moments would allow for a larger class of 
functions and remains an open problem. Additionally, handling discontinuous functions, which 
typically arise as conditional statements in Monte Carlo programs, requires new techniques and 
assumptions beyond our current framework.